\documentclass[12pt,authoryear]{elsarticle}

\usepackage{amssymb}
\usepackage{amsmath}

\usepackage[linesnumbered,ruled,vlined]{algorithm2e} 
\usepackage{xcolor}
\usepackage{enumerate}
\usepackage{hyperref}

\usepackage{multirow}
\usepackage{longtable}

\usepackage{subcaption}

\journal{Computers and Operations Research}

\begin{document}

\begin{frontmatter}



\title{Random-Key Optimizer and Linearization for the Quadratic Multiple Constraints Variable-Sized Bin Packing Problem}


\author[label1]{Natalia Alves Santos\corref{cor1}} 
\cortext[cor1]{Corresponding author.}
\ead{natalia.santos26@unifesp.br}

\author[label1,ita]{Marlon Jeske} 
\ead{jeske@unifesp.br}

\author[label1]{Antônio Augusto Chaves} 
\ead{antonio.chaves@unifesp.br}

\affiliation[label1]{organization={Federal University of São Paulo},
            addressline={Av. Cesare Mansueto Giulio Lattes, 1201}, 
            city={São José dos Campos},
            postcode={12247-014}, 
            state={SP},
            country={Brazil}}

\affiliation[ita]{organization={Aeronautics Institute of Technology},
            addressline={Praça Marechal Eduardo Gomes, 50}, 
            city={São José dos Campos},
            postcode={12228-900}, 
            state={SP},
            country={Brazil}}

\begin{abstract}
This paper addresses the \textit{Quadratic Multiple Constraints Variable-Sized Bin Packing Problem} (QMC-VSBPP), a challenging combinatorial optimization problem that generalizes the classical bin packing problem by incorporating multiple capacity dimensions, heterogeneous bin types, and quadratic interaction costs between items. We propose two complementary methods that advance the current state-of-the-art. First, a linearized mathematical model is introduced to eliminate quadratic terms, enabling the use of exact solvers such as Gurobi to compute strong lower bounds—reported here for the first time for this problem. Second, we develop RKO-ACO, a continuous-domain Ant Colony Optimization algorithm within the Random-Key Optimizer framework, enhanced with adaptive Q-learning parameter control and efficient local search. Extensive computational experiments on benchmark instances show that the proposed linearized model produces significantly tighter lower bounds than the original quadratic model, while RKO-ACO consistently matches or improves upon all best-known solutions in the literature, establishing new upper bounds for large-scale instances. These results provide new reference values for future studies and demonstrate the effectiveness of evolutionary and random-key approaches for solving complex quadratic packing problems.
\\
\\
\textit{Published on Computers and Operations Research. }\url{https://doi.org/10.1016/j.cor.2026.107482}
\end{abstract}


 
    
    



\end{frontmatter}



\section{Introduction} \label{sec:intro}

The bin packing problem (BPP) is a well-studied combinatorial optimization problem that has been extensively researched since its introduction in 1979 and has been applied in various fields, such as scheduling and cutting stock problems. The primary objective of the BPP is to minimize the packing cost of a set of items into a finite number of bins, each with a fixed capacity, while considering the weights of the items. 

The variable-sized bin packing problem (VSBPP), one of many variants of BPP, allows packing combinations in different types of bins, each with its own cost and capacity. Besides, BPPs and VSBPPs with conflicts often penalize when certain pairs of items are assigned to the same bin \citep{Ekici2023}, or offer rewards for items of the same category packed together \citep{Santos2019}. Recently, a novel variant of the VSBPP was introduced, known as the Quadratic Multiple Constraint VSBPP (QMC-VSBPP) \citep{Meng2022}. This variant considers multiple dimensions for bin capacities and item weights, and penalizes pairs of items packed in different bins. As the VSBPP, the QMC-VSBPP is classified as NP-hard, indicating its computational complexity.

To address the inherent complexity of the QMC-VSBPP, this work utilizes the Random-Key Optimizer (RKO) framework \citep{Chaves2025_RKOpaper}. By operating in a problem-agnostic continuous domain, RKO decouples the metaheuristic search engine from problem-specific constraints, allowing for the integration of continuous-space metaheuristics, such as the \textit{Ant Colony Optimization for Continuous Domains} (ACO) \citep{Dorigo2008_ACO}.

In this work, we address the QMC-VSBPP, aiming to improve both solution quality and the quality of lower bounds. To this end, we formulate three core research questions: 
(1) how to linearize the model of the QMC-VSBPP to achieve improved lower bounds and enhance the solution quality; (2) to explore whether the RKO framework, incorporating a continuous-domain ACO algorithm, can successfully generate high-quality solutions for this problem; and (3) to compare the efficiency of the solutions obtained through the proposed approach against existing benchmarks. Based on these questions, we hypothesize that the proposed linearization of the QMC-VSBPP model will reduce the solution gap and yield tighter lower bounds, consequently increasing the performance of standard solvers like Gurobi. Furthermore, we tailored the RKO framework with the continuous-domain ACO, which will produce superior solutions compared to traditional methods found in the literature, leading to the establishment of new best-known solutions for the QMC-VSBPP.


The main contributions of this work are:

\begin{itemize}

    \item Advances the state-of-the-art for the QMC-VSBPP with exact and metaheuristic methods.
    
    \item Proposes a new linearized model, eliminating complex quadratic terms.
    
    \item Introduces lower bounds via exact solutions of the linearized model using Gurobi.
    
    \item Develops the RKO-ACO: an adaptation of continuous ACO for the Random-Key Optimizer.
    
    \item RKO-ACO found new best-known solutions compared to those previously reported in the literature.

\end{itemize}

This paper is organized as follows. Section~\ref{sec:lit_review} reviews the literature on the bin packing problem with conflicts, bio-inspired heuristics, and RKO. Section~\ref{sec:problem} presents the model formulation, and the proposed linearization. Section~\ref{sec:theory} provides the theoretical background on the Random-Key Optimizer and ACO for continuous domains and introduces the developed RKO-ACO algorithm. Section~\ref{sec:methodology} details the implementation of RKO-ACO tailored to the QMC-VSBPP, with a focus on the solution decoder. Section~\ref{sec:results} presents computational experiments and analyzes the results obtained from exact and metaheuristic approaches. Section~\ref{sec:conclusion} concludes with the main contributions, addresses research questions and validates the hypotheses.

\section{Literature Review} \label{sec:lit_review}

The bin packing problem and its variants, including the variable-sized bin packing problem, have been extensively studied over the past decades. This review is organized into three subjects: BPP with conflicts, bio-inspired metaheuristics for BPP, and the RKO framework and its applications.

Several extensions of BPP incorporate conflict constraints, where specific pairs of items cannot be packed together or incur penalties when placed in separate bins. Early studies on the classical BPP with conflicts (BPC) proposed the first-fit decrease (FFD) heuristic adaptations and clique computations on the conflict graph \citep{Gendreau2004}. Later works improved the results using set covering formulations with column generation and tabu search \citep{Muritiba2010}. Online variants, where items are partially or completely unknown a priori, were also explored \citep{Epstein2011}. The min-conflict packing problem \citep{Khanafer2012} introduced a bi-objective formulation minimizing both violated conflicts and the number of bins. Overall, the main challenge lies in resolving the conflict graph rather than the packing constraints themselves \citep{Khanafer2012}.

Variants that consider item fragmentation (BPPC-IF) allow items to be partially packed while respecting conflicts. Heuristics for sequential packing based on item degree in the conflict graph were proposed \citep{Ekici2021}, and later improved by refining item selection criteria \citep{Fleszar2022}. The density of the conflict graph strongly influences the solution gap, with higher density typically leading to worse outcomes. Other related variants include packing compatible categories \citep{Santos2019} and Open-End BPC, where the last item may exceed bin capacity \citep{Balik2025}. These studies commonly employed Variable Neighborhood Search (VNS) with initial solutions generated by modified FFD heuristics.

Recent developments include the VSBPP with conflicts (VSBPPC) and item fragmentation \citep{Ekici2022}, where maximal clique calculations provide lower bounds, and heuristics generate independent subsets of compatible items to guide packing. In \citet{Ekici2023}, the VSBPPC was formulated as a mixed-integer linear programming (MILP) problem and solved using a Large Neighborhood Search (LNS) metaheuristic. Firstly, initial solutions are constructed using a greedy heuristic by sequentially packing items into existing or new bins while minimizing cost. Then, the LNS procedure iteratively destroys a subset of bins and reconstructs them using the initial solution strategy combined with local search operations, such as moving or swapping items, and changing bin types to smaller costs as much as possible.

In \citet{Meng2022}, the VSBPPC was extended to the Quadratic Multiple-Constraint VSBPP (QMC-VSBPP), introducing 3 to 5 attribute dimensions, where each item has a $d$-dimensional weight vector and each bin has corresponding $d$-dimensional capacity limits. This formulation models scenarios such as cloud computing resource allocation, where dimensions represent attributes like CPU and RAM. In addition to capacity constraints, the conflict graph imposes extra costs when specific item pairs are placed in different bins. The authors formulated the problem as an MILP and proposed a VNS algorithm, with initial solutions generated via hierarchical clustering. The VNS employs neighborhood strategies including bin swapping, bin deletion, and shortest-path-based item permutation. Performance was evaluated against adapted VNS and Genetic Algorithm methods from the VSBPP state-of-the-art literature \citep{Haouari2009, Hemmelmayr2012} as well as the CPLEX solver. Across 96 newly generated benchmark instances, the proposed VNS consistently found better solutions in less computational time than the other evaluated approaches.

Bio-inspired optimization methods have been widely applied to BPP and its variants. Approaches include squirrel search algorithms \citep{Ashmawi2019}, evolutionary heuristics \citep{Stawowy2008}, and genetic algorithms using grouping strategies and parallel islands \citep{Kucukyilmaz2018}. Particle Swarm Optimization (PSO) has been explored for multi-objective BPPs \citep{Liu2008}, while Ant Colony Optimization (ACO) has been applied in classic form with local search \citep{Levine2004_ACO}, as well as enhanced with differential pheromone strategies \citep{Ali2024_ACO}. Moreover, \citet{Dorigo2008_ACO} demonstrated that ACO, when adapted for continuous domains, achieves superior performance across a range of combinatorial problems, highlighting its potential for mixed-variable optimization challenges. 

The concept of random keys was introduced by \citet{bean1994_randomkeys} to facilitate search over continuous spaces in combinatorial problems. As a precursor to the RKO framework, \citet{Chaves2024_RK_GRASP} adapted GRASP for random-key optimization, demonstrating effectiveness across several NP-hard problems. More recently, the RKO framework has been successfully applied to classical optimization problems, including the Traveling Salesman Problem, Set Covering, Vehicle Routing, and Node Capacitated Graph Partitioning \citep{Chaves2025_RKOpaper}, as well as real-world applications such as robot motion planning \citep{Schuetz2022},  operating room scheduling \citep{Vieira2025}, and cutting stock problems \citep{Silva2025}.

The reviewed studies demonstrate the versatility of bin packing extensions, the effectiveness of bio-inspired metaheuristics, and the adaptability of the RKO framework to complex combinatorial problems. However, to the best of our knowledge, the QMC-VSBPP has not been explored in the literature beyond its formulation and instance generation. This gap provides the foundation for our work, which seeks to advance exact and metaheuristic approaches for this challenging variant.

\section{Problem Definition} \label{sec:problem}

\begin{figure}[htb]
    \centering
    \includegraphics[width=1\linewidth]{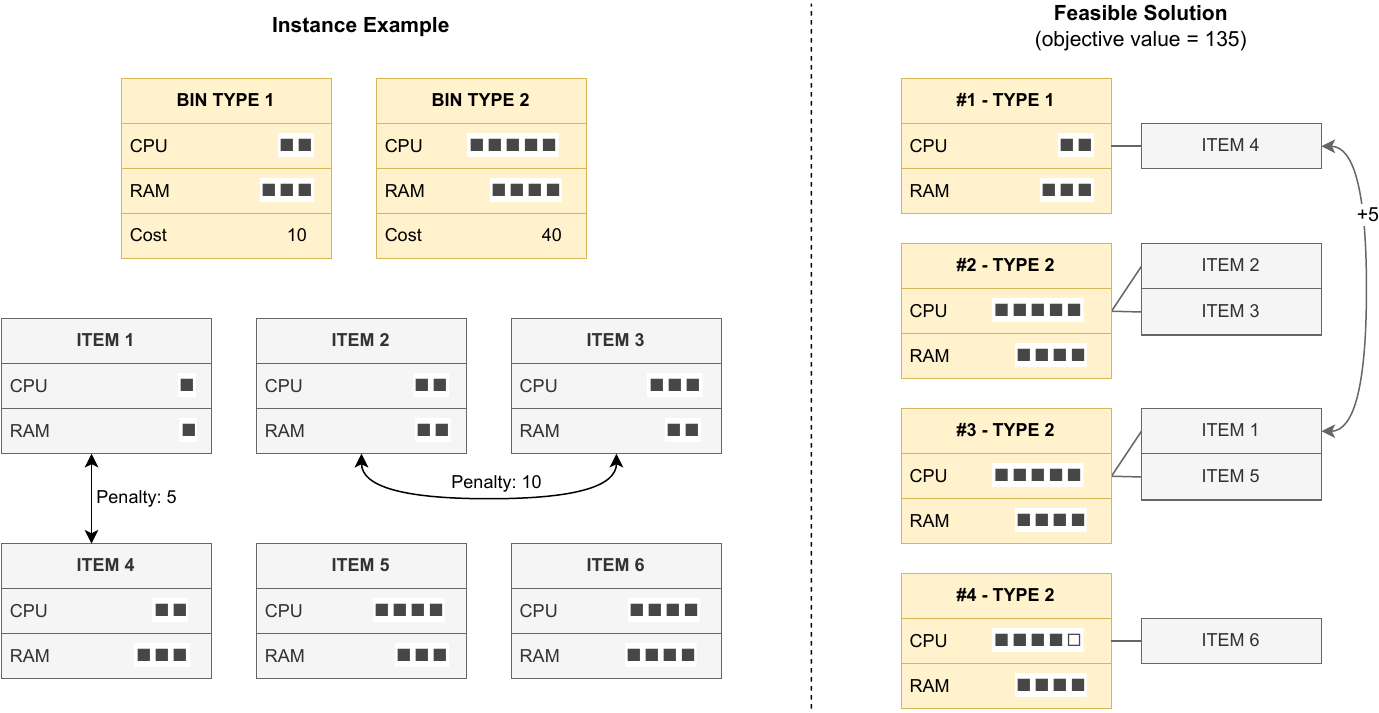}
    \caption{Instance example}
    \label{fig:instance}
\end{figure}

The Quadratic Multiple-Constraint Variable-Sized Bin Packing Problem (QMC-VSBPP) proposed by \citet{Meng2022} incorporates multiple capacity dimensions and quadratic interaction costs. Unlike the traditional VSBPP, which models a single ``size'' dimension, this variant assigns $d$-dimensional weight vectors to items and bins to represent cloud computing resources such as CPU and RAM. A feasibility requirement is that an item's weight must not exceed the available bin capacity in any of the $d$ dimensions. For example, a resource (item) requiring 32 GB of RAM cannot be allocated to a server (bin) with only 4 GB of capacity, regardless of whether other resource requirements are met. Additionally, penalties are incurred whenever specific pairs of items are placed in separate bins, modeling communication latency costs between interdependent resources.
Figure \ref{fig:instance} illustrates an instance with 6 items and 2 bins, demonstrating the challenge of selecting optimal bin types for cost efficiency while minimizing allocation conflicts.

\subsection{Mathematical Model}


The QMC-VSBPP formulated by \citet{Meng2022} involves a set of items \( I = \{1, 2, \ldots, n\} \), a collection of bin types \( M = \{1, 2, \ldots, m\} \), and multiple attribute dimensions \( D = \{1, 2, \ldots, d\} \).

Each item \( i \in I \) is characterized by a weight vector \( \mathbf{w}_i = (w_{i1}, w_{i2}, \ldots, w_{id}) \), where \( w_{ir} > 0 \) denotes the weight of item \( i \) in dimension \( r \in D \). For any distinct pair of items \( i, s \in I \), a non-negative joint cost \( c_{is} \) with \(1 \leq i \neq s \leq n\) is incurred if the items are placed in different bins.

Each bin type \( k \in M \) is associated with a fixed cost \( c_k \) and a capacity vector \( \mathbf{Q}_k = (Q_{k1}, Q_{k2}, \ldots, Q_{kd}) \), where \( Q_{kr} > 0 \) denotes the capacity in dimension \( r \in D \). Bins can be instantiated in unlimited numbers, with their types chosen from \( M \).

Consider \( B = \{1, 2, \dots, n\} \) as the set of potential bins, with $n$ as the maximum number of bins considering the assignment of one item per bin. The binary variable \( x_{ij} \) is equal to 1 if item \( i \) is assigned to bin \( j \in B\), and 0 otherwise. Also, \( y_{jk} \) equals 1 if bin \( j \in B \) is assigned to the type \( k \in M\), and 0 otherwise.

The problem objective is to allocate all items into bins satisfying the conditions: the total weight of items in any bin does not exceed its capacity in any dimension; each opened bin must be assigned to a specific bin type; and both incurring costs, the cost per bin type utilized and the penalty cost for packing certain items in separate bins, must be minimized.

The mathematical model for the QMC-VSBPP is expressed as follows \citep{Meng2022}:

\begin{align}
    \min & \sum_{j \in B} \sum_{k \in M} c_k y_{jk} + \sum_{j \in B} \sum_{i \in I} \sum_{s \in I} c_{is} x_{ij}(1 - x_{sj}) \label{qmc_fo}
\end{align}
\begin{align}
    & \sum_{j \in B} x_{ij} = 1 && \forall i \in I \label{qmc_c1} \\
    & \sum_{k \in M} y_{jk} \leq 1 && \forall j \in B \label{qmc_c2} \\
    & \sum_{i \in I} w_{ir} x_{ij} \leq \sum_{k \in M} Q_{kr} y_{jk} && \forall j \in B,\; r \in D \label{qmc_c3} \\
    & x_{ij} \in \{0,1\} && \forall i \in I,\; j \in B \label{qmc_c4} \\
    & y_{jk} \in \{0,1\} && \forall j \in B,\; k \in M. \label{qmc_c5}
\end{align}

The first term in the objective function \eqref{qmc_fo} sums the costs of the bins utilized, while the second term represents the penalty costs by placing certain items in separate bins. Constraint \eqref{qmc_c1} ensures that each item is assigned to only one bin. Constraint \eqref{qmc_c2} ensures that each opened bin is assigned to at most one type. Constraint \eqref{qmc_c3} ensures that the total weight of items assigned to a bin does not exceed the capacity of its assigned type in any dimension. Finally, constraints \eqref{qmc_c4} and \eqref{qmc_c5} set the decision variables' binary domain. For example, the objective function value of the solution shown in Figure \ref{fig:instance} is 135.

\subsection{Model Linearization} \label{sec:linear}

Linearization is a common approach used to simplify optimization problems with quadratic terms, reducing computational complexity and improving the efficiency of solvers in finding optimal solutions.

For the QMC-VSBPP model linearization, a new set of binary variables, $z_{ijs}$, is introduced to capture the interaction between items $i$ and $s$ when they are placed in different bins $j$, in substitution of the objective function's quadratic term $x_{ij}(1 - x_{sj})$. The linearized model is defined as follows:

\begin{align}
    \min & \sum_{j \in B} \sum_{k \in M} c_k y_{jk} + \sum_{j \in B} \sum_{i \in I} \sum_{s \in I} c_{is} z_{ijs} \label{qmcl_fo}
\end{align}
\begin{align}
    & \sum_{j \in B} x_{ij} = 1 && \forall i \in I \label{qmcl_c1} \\
    & \sum_{k \in M} y_{jk} \leq 1 && \forall j \in B \label{qmcl_c2} \\
    & \sum_{i \in I} w_{ir} x_{ij} \leq \sum_{k \in M} Q_{kr} y_{jk} && \forall j \in B,\; r \in D \label{qmcl_c3} \\
    & z_{ijs} \leq x_{ij} && \forall i \in I, j \in B,\; s \in I \label{qmcl_c4} \\
    & z_{ijs} \leq 1- x_{sj} && \forall i \in I, j \in B,\; s \in I \label{qmcl_c5} \\
    & z_{ijs} \geq x_{ij} - x_{sj} && \forall i \in I, j \in B,\; s \in I \label{qmcl_c6} \\
    & x_{ij} \in \{0,1\} && \forall i \in I,\; j \in B \label{qmcl_c7} \\
    & y_{jk} \in \{0,1\} && \forall j \in B,\; k \in M \label{qmcl_c8} \\
    & z_{ijs} \in \{0,1\} && \forall i \in I, j \in B,\; s \in I. \label{qmcl_c9}
\end{align}

The new constraints involving $z_{ijs}$ ensure that these variables correctly represent the interaction between items. Constraints \eqref{qmcl_c4}-\eqref{qmcl_c6} refer to the model linearization. Constraint \eqref{qmcl_c9} defines the $z_{ijs}$ as binary.

\section{Random-Key Optimizer (RKO)} \label{sec:theory}

A typical combinatorial optimization problem consists of a finite ground set $E=\{1, \ldots,n\}$, an objective function $f: 2^E \rightarrow \mathbb{R} $ , and a set of feasible solutions $F \subseteq 2^E$. For minimization problems, the goal is to find an optimal solution $S^* \in F$ such that $f(S^*) \leq f(S) \quad \forall S \in F$. 
Within the random keys approach, a problem solution can be encoded as a vector $\chi$ of real numbers, where $\chi = (x_1,x_2, \ldots,x_n)$ and $x_i \in [0,1)$. Thus, $\chi$ is a point in the unit hypercube $[0,1)^n \subset \mathbb{R}^n$, allowing the search process to operate over a continuous domain instead of the discrete solution space originally defined by the problem. To convert a random-key vector into a problem solution, a decoder function $G$ is employed to map $\chi$ into a problem-specific feasible solution $S \in F$, i.e., $S = G(\chi)$ (where $G:[0,1)^n \to F$).

\subsection{RKO Framework}

The RKO framework, proposed by \cite{Chaves2025_RKOpaper}\footnote{\url{https://github.com/RKO-solver}}, provides a modular environment in which metaheuristics search directly in the continuous random-key space, while a dedicated decoding function maps each random-key vector into a feasible discrete solution. This clear separation of search and problem-specific logic makes algorithms built within the RKO framework largely problem-independent: metaheuristic operators explore the continuous space, and the decoding function handles all problem-specific constraints and requirements.

In summary, each iteration follows a standardized cycle:

\begin{enumerate}
\item Random key generation: initialize an individual or population of random keys.

\item Solution decoding: map each random key to a problem solution, evaluate it, and optionally apply local search in the problem domain.

\item Continuous-space exploration: apply continuous-space search methods, such as the Nelder–Mead algorithm \citep{Chaves2024_RK_GRASP}, to generate new random keys.
\end{enumerate}

To enhance performance, the framework integrates several reusable, problem-independent strategies. Multi-threading allows each processing thread to run an independent metaheuristic instance, whether multiple runs of the same algorithm or different algorithms in parallel. A shared solution pool collects candidate solutions from all threads, creating a set of elite solutions. A restart mechanism helps escape stagnation by fully reinitializing both the population and the solution pool. Finally, an online parameter tuning component, based on Q-learning, dynamically adjusts search parameters to balance intensification and diversification over time.

These strategies, along with the continuous random-key search design, make the framework compatible with any metaheuristic able to operate in continuous space \citep{Chaves2025_RKOpaper}.
In this work, the RKO is extended with a continuous-domain ACO metaheuristic, hereafter referred to as RKO-ACO, as presented in the next sections.

\subsection{ACO for Continuous Domains} \label{sec:aco}

Ant Colony Optimization is a well-established metaheuristic inspired by the foraging behavior of ants, traditionally applied to discrete combinatorial problems. To address continuous optimization challenges, the ACO paradigm has been extended to operate in continuous domains \citep{Dorigo2008_ACO}. In this variant, the pheromone model is represented by an archive of elite solutions, each encoded as a vector in the continuous search space.

In the continuous ACO approach, each solution in the archive is assigned a weight reflecting its quality and rank. New candidate solutions are generated by probabilistically selecting an archive member according to these weights and then sampling each variable from a Gaussian distribution centered at the corresponding value in the chosen solution. The standard deviation of this Gaussian is adaptively determined by the dispersion of the archive in each dimension, scaled by a parameter $\xi$.

Formally, let $\kappa$ denote the archive size, $q$ the selection pressure parameter, and $\xi$ the exploration scaling parameter. The solutions in the archive are sorted according to their objective function values, denoted as $S_1, S_2, \ldots, S_\kappa$, such that $f(S_1) \leq f(S_2) \leq \ldots \leq f(S_\kappa)$. The weight $\omega_l$ for the $l$-th ranked solution is computed as:

\begin{equation}
\omega_l = \frac{1}{q\kappa\sqrt{2\pi}} \exp\left(-\frac{(l-1)^2}{2q^2\kappa^2}\right),
\end{equation}

\noindent where $l=1$ corresponds to the best solution ($S_1$). The probability of selecting a solution is proportional to its weight.

For each variable $i$ in the random-key vector, the standard deviation $\sigma_l^i$ for sampling is given by:

\begin{equation}
\sigma_l^i = \xi \frac{1}{\kappa-1} \sum_{e \neq l} \left| x_e^i - x_l^i \right|,
\end{equation}

\noindent where $x_l^i$ denotes the value of the $i$-th variable of the $l$-th ranked solution, and the sum is over all other archive members. In cases where the computed $\sigma_l^i$ is numerically close to zero (i.e., negligible dispersion), a large default $\sigma_l^i$ (e.g., 0.9999) is used to preserve exploration.

Therefore, new solutions are created as follows:

\begin{enumerate}
    \item Compute weights for all archive solutions and derive their selection probabilities.
    \item Select an archive solution according to these probabilities.
    \item For each variable, sample a new value from a Gaussian distribution centered at the selected solution's value, with standard deviation as above. Rejection sampling is used to ensure that variables remain within the feasible range $[0,1)$.
\end{enumerate}

This process is iterated for a specified number of ants (number of candidate solutions) per generation. The archive is then updated by merging the new solutions with the existing archive and retaining only the top $k$ solutions according to fitness.

\subsection{RKO-ACO}

The proposed RKO-ACO metaheuristic integrates a continuous Ant Colony Optimization approach, as detailed in the previous section, with Q-learning for dynamic parameter adaptation and the Nelder–Mead algorithm for local search. The algorithm maintains an archive of elite solutions that guides the search and a global solution pool shared across multi-threaded executions of the RKO-ACO metaheuristic. 

\begin{figure}[!htb]
    \centering
    \includegraphics[width=1\linewidth]{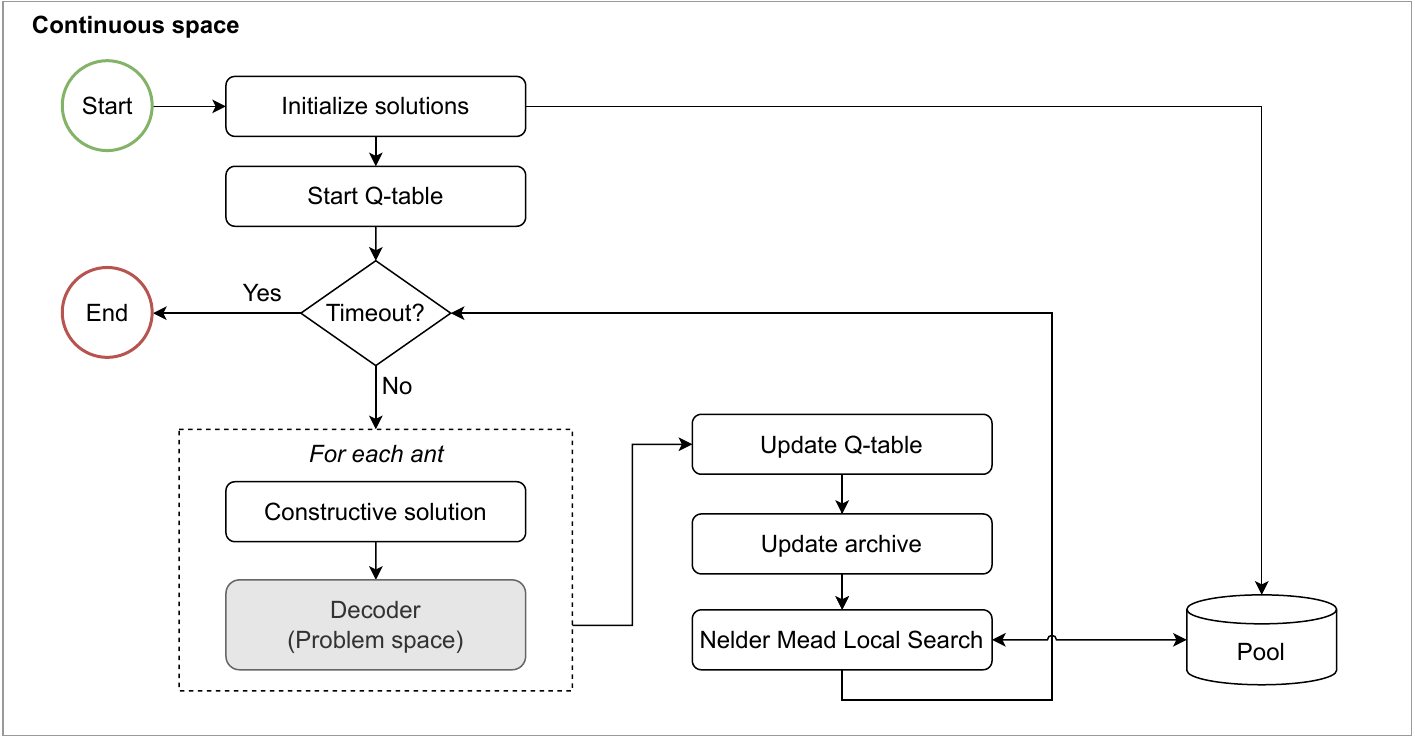}
    \caption{RKO-ACO flowchart}
    \label{fig:flowchart}
\end{figure}

The step-by-step procedure of the algorithm is illustrated in Figure~\ref{fig:flowchart}.

In summary, the RKO-ACO operates as follows:

\begin{enumerate}
\item \textbf{Solution Initialization:} The solution archive is initialized as a randomized random-key, or using a constructive heuristic tailored to the problem (as detailed in Section \ref{subsec:initial_solutions}). Each solution is encoded as a random-key vector, and the best initial solution is submitted to the global solution pool.

\item \textbf{Q-table Initialization:} The algorithm initializes its parameters within predefined value ranges, delimiting the space explored by the Q-learning agent during the optimization process. The Q-learning process is explained in Section \ref{subsec:qlearning}.

\item \textbf{Generations Loop:}
\begin{itemize}

\item \emph{Solution Construction:} Each ant constructs a new solution, as described in Section~\ref{sec:aco}, resulting in a random-key. 

\item \emph{Solution Decoder:} The constructed random-key vector is decoded into a feasible solution in the problem-space, as described in Section \ref{sec:methodology}. Here, its objective value (fitness) is evaluated.

\item \emph{Q-table Update:} The Q-learning agent receives a reward based on the relative improvement in solution quality. This feedback guides the selection of parameter values and reinforces combinations that lead to better performance.

\item \emph{Archive Update:} The archive is updated with the newly generated solutions and pruned to retain only the best individuals, as detailed in Section \ref{sec:aco}.

\item \emph{Local Search:} The best solution of the current generation undergoes local improvement using the Nelder–Mead algorithm.

\item \emph{Solution Pool Update:} When a new best solution is found, it is added to the shared solution pool, which retains the global best solutions across all threads.

\end{itemize}

\item \textbf{Termination and Restart:} The process repeats until a predefined computational time limit is reached. If a restart condition is triggered, the solution pool is cleared, and new solutions are initialized, restarting the ACO generation process to escape stagnation.
\end{enumerate}

A detailed description of RKO-ACO's online parameter tuning and a novel caching system for the solution decode step are presented in the following subsections.

\subsubsection{Q-learning for Online Parameter Tuning}
\label{subsec:qlearning}

To enhance metaheuristic performance, ACO parameters are dynamically adapted during execution using a Q-learning approach. The parameters $\kappa$, number of ants, $q$, and $\xi$ are randomly initialized within predefined ranges in a Q-table. These parameters are subsequently updated at each generation based on a reward function, presented in \citet{Chaves2025_RKOpaper}, that evaluates improvements in solution quality, favoring configurations that lead to better solutions.

For the archive size parameter, changes between generations require careful management. When $\kappa$ decreases, the worst-performing solutions, ranked by objective function value, are removed. Conversely, when $\kappa$ increases, additional solutions are created by sampling new random-key vectors with values uniformly distributed in $[0,1]$. Each vector is decoded into a problem solution, and this process continues until the new archive capacity is reached. 

This adaptive mechanism manages the trade-off between diversity and efficiency: larger $\kappa$ values enhance exploration by maintaining a more diverse population, whereas frequent or substantial expansions increase computational cost due to the evaluation of newly generated solutions.

\subsubsection{Caching}

As a contribution to the RKO framework, we implemented a fixed-size queue-based caching system to avoid recalculating solutions that have already been evaluated on the Decode step. The cache stores previously evaluated solutions as key-value pairs, where the key is a string representation of the solution and the value is its corresponding objective function value. The cache maintains a maximum size ($Q_{\max} = 1000$); when the limit is reached, the oldest entry (the one inserted earliest) is removed to accommodate new solutions.
The caching process is presented in Algorithm~\ref{alg:caching}.

\begin{algorithm}[!htb]
    \footnotesize
    \caption{Queue-Based Caching}
    \label{alg:caching}
    \KwIn{Decoded solution $S$, objective value $f(S)$ (optional), cache $\mathcal{Q}$ of max size $Q_{\max}$}
    
    \If{$S$ exists in $\mathcal{Q}$}{
        \Return{cached f(S) for $S$}
    }
    \Else{
        \If{size of $\mathcal{Q} = Q_{\max}$}{
            Remove the entry at the front of the queue from $\mathcal{Q}$\;
        }
        Insert $(S, f(S))$ at the back of the queue into $\mathcal{Q}$\;
        \Return{$f(S)$}
    }
    \end{algorithm}

Although the cache may introduce additional overhead at the beginning of the metaheuristic execution, when most solutions are new, it helps reduce computation time during convergence by skipping repeated evaluations of known solutions.

\section{RKO-ACO for QMC-VSBPP} \label{sec:methodology}


This section details the adaptation of the problem-agnostic RKO-ACO framework to solve the QMC-VSBPP. As RKO-ACO operates on random-key vectors, a problem-specific \textit{solution decoder} is required to map the continuous search space to the discrete problem space. We explain this decoding mechanism in detail, along with the generation of initial solutions, post-processing steps, local search procedure within the problem space, and tie-breaking criteria for the objective function.

\subsection{Notations}

Throughout this work, we use the following notations, which are fundamental to the description of our algorithms and analysis.
 
The adjacency matrix $L = [\ell_{is}]$, also called the \emph{links}, encodes the pairwise penalties for separating items. Specifically, $\ell_{is} = c_{is}$ if items $i$ and $s$ can feasibly be packed together in at least one bin type (i.e., their combined weights do not exceed the capacity of the largest bin type in any dimension). 

The \emph{aggregate neighbor penalty} of an item $i$ is denoted by $\ell_i^+$ and defined as follows:
\[
\ell_i^+ = \sum_{s=1}^n \ell_{is}   \,\, \forall i \in I,
\]
where $n$ is the number of items and $\ell_{is}$ is the penalty (link cost) for separating items $i$ and $s$.

The \emph{aggregate neighbor weight} of an item $i$ is denoted by $w_i^+$ and defined as follows:
\[
w_i^+ = \sum_{s=1}^n \mathbb{I}[\ell_{is} > 0] \left( \sum_{r=1}^d w_{sr} \right) \,\, \forall i \in I,
\]
where $w_{sr}$ is the weight of item $s$ in dimension $r$, and $\mathbb{I}[\ell_{is} > 0]$ is 1 if items $i$ and $s$ are linked, and 0 otherwise.

The \emph{largest bin type}, denoted $k^*$, is defined as the bin type with the largest aggregated capacity:
\[
k^* = \arg\max_{k \in M} \left( \sum_{r=1}^d Q_{kr} \right),
\]

Lastly, consider that the bin types are sorted in ascending order of their cost.

\subsection{Generating Initial Solutions} 
\label{subsec:initial_solutions}

Initial solutions are constructed using a semi-greedy randomized procedure that prioritizes the assignment of items with high aggregate neighbor weight \(w_i^+\) while maintaining feasibility. At each iteration, a randomized candidate list (RCL) of unassigned items with links is formed, and the item with the highest \(w_i^+\) in the RCL is selected as the starting point for a new bin of type \(k^*\). Additional items are greedily added to the bin based on the highest penalty \(\ell_{is}\) to the start item, case they fit within the bin's capacity constraints. This process repeats until all items with links are assigned. Any remaining unassigned items are placed in the existing bin that minimizes the link cost and can accommodate the item, or in a new bin if necessary. The complete procedure, including post-processing steps, is detailed in Algorithm~\ref{alg:semi-greedy-init}.

\begin{algorithm}[!htb]
    \footnotesize
    \caption{Semi-Greedy Initial Solution Generation}
    \label{alg:semi-greedy-init}
    Initialize all items as unassigned\;
    Initialize an empty set of opened bins $B$\;
    Randomly select RCL size $\tau$ uniformly between 3\% and 5\% of $n$\;
    \While{there are unassigned items with $\ell_{i}^+ > 0$}{
        Build RCL: take up to $\tau$ unassigned items in input order\;
        $i^* \gets \arg \max_{i \in \text{RCL}} w_i^+$\;
        Open a new bin $j \in B$ of type $k^*$. Assign $i^*$ to $j$\;
        Mark $i^*$ as assigned\;
        \While{true}{
            Among all unassigned items $s$, find $s^*$ with the highest link cost $\ell_{i^*s}$ such that $s^*$ fits in the current bin $j$\;
            \If{$s^*$ exists}{
                Assign $s^*$ to $j$ and mark as assigned\;
            }
        }
    }
    \For{each remaining unassigned item $s$}{
        Assign $s$ to a bin in $B$ that minimizes link cost and fits, or  open a new bin (Algorithm~\ref{alg:assign_best})\;
    }
    Apply post-processing (bin type replacement, bin merging)\;
    Apply local search with probability $p=1$\;
    Encode the final allocation as a random-key vector $\chi$\;
    Compute the objective value\;
    \Return{$\chi$}
\end{algorithm}

\begin{algorithm}[!htb]
    \footnotesize
    \caption{Item to Best Bin Assignment}
    \label{alg:assign_best}
    \KwIn{Opened bins $B$, unassigned item $i$}
    $B^*$ $\gets$ \{ $j \in B$ : $i$ fits in $j$ \}\;
    \If{$B^* \neq \emptyset$}{
        $j^* \gets \arg\min_{j \in B^*} \sum_{s \in I(j)} \ell_{is}$\;
        Assign $i$ to $j^*$\;
    }
    \Else{
        Open new bin $j$ of type $k^*$. Assign $i$ to $j$\;
    }
    \Return{Item assignment}
\end{algorithm}

\subsection{Decoding and Item to Bin Assignment}

Each solution is represented as a random-key vector of length $n+3$. The first $n$ elements determine the allocation order: items are sorted in ascending order of their random-key values, and this order is used for assignment. The last three elements of the vector encode algorithmic parameters: (i) the allocation strategy, (ii) the number of initial bins to open, and (iii) the probability of applying item relocation during local search.

We propose three item-to-bin assignment strategies. The first reuses the semi-greedy procedure for building initial solutions, with the order of unassigned items following the decoded random-key vector. The second and third strategies are adapted from the random-key decoding method for the node-capacitated graph partitioning problem (NCGPP) \citep{Chaves2025_RKOpaper}. In these approaches, items are processed in the decoded order: each item is assigned to the best bin that can accommodate it without exceeding capacity, or in a new bin if none are suitable. The second strategy uses a random-key parameter to define the number of bins to pre-open, whereas the third starts with no bins opened a priori. The Algorithm \ref{alg:decoding} describes the assignment process.

\begin{algorithm}[!htb]
    \footnotesize
    \caption{Random-key Decode and Item Assignment}
    \label{alg:decoding}
    \KwIn{Random-key vector $\chi$ of length $n+3$}
    
    Sort the first $n$ elements of $\chi$ in ascending order to obtain item order $\gamma$\;
    Extract allocation strategy, number of initial bins $\nu$, and item relocation probability $p$ from the last three elements of $\chi$\;
    
    \uIf{$strategy=1$}{
        Apply the semi-greedy construction (Algorithm~\ref{alg:semi-greedy-init}) using $\gamma$\;
    }
    \uElse{
        \If{$strategy=2$ and $\nu > 0$}{
            Open $\nu$ bins of type $k^*$\;
            Assign the first items in $\gamma$ to these bins (one per bin)\;
        }

        \For{each unassigned item $i$ in $\gamma$ with $\ell_i^+>0$}{
            Assign $i$ to the best feasible bin (Algorithm~\ref{alg:assign_best})\;
        }
    }
                
    \For{each remaining unassigned item $i$ in $\gamma$}{
        Assign $i$ to the best feasible bin (Algorithm~\ref{alg:assign_best})\;
    }
    
    Apply post-processing (bin type replacement, bin merging)\;
    Apply local search with probability $p$\;
    \Return{Set of opened bins $B$ with its assigned items}
\end{algorithm}

\subsection{Post-processing}

Post-processing is applied both to initial solutions and to every decoded solution to improve cost efficiency while maintaining feasibility. Two main procedures are performed.

The \textit{bin type replacement} checks if any bin can be replaced with a smallest-cost bin type that remains feasible for its assigned items.

After, the \textit{bin merging} verifies for every pair of opened bins if the combined contents of two bins fit within a feasible bin type whose cost is lower than the total cost of the original bins. Thus, the items are reassigned to that bin, and the redundant bin is removed.

The following Algorithms \ref{alg:bin_type_replacement} and \ref{alg:bin_merging} describe these procedures in detail.

\begin{algorithm}[!htb]
    \footnotesize
    \caption{Bin Type Replacement}
    \label{alg:bin_type_replacement}
    \KwIn{Set of opened bins $B$}
    \For{each bin $j \in B$}{
        \For{each bin type $k$ inferior than current type of $j$}{
            \If{all items in $j$ fit in bin type $k$}{
                Update $j$ to type $k$\;
                \textbf{break}\;
            }
        }
    }
    \Return{Updated $B$}
\end{algorithm}

\begin{algorithm}[!htb]
    \footnotesize
    \caption{Bin Merging}
    \label{alg:bin_merging}
    \KwIn{Set of opened bins $B$}
    
    Initialize set $T$ to track merged bins\;
    
    \For{each bin $j \in B$}{
        \If{$j \in T$}{
            \textbf{continue}
        }
        \For{each bin $j' \in B$}{
            \If{$j' \in T$}{
                \textbf{continue}
            }
            \For{each bin type $k$ with cost $c_k \leq c_j + c_{j'}$ }{
                \If{all items in $j \cup j'$ fit in bin type $k$}{
                    Create new bin $j''$ in $B$ of type $k$\;
                    Move items from $j$ and $j'$ to $j''$\;
                    Add $j$ and $j'$ to $T$\;
                    \textbf{break} (only merge each pair once)\;
                }
            }
            \If{$j \in T$}{
                \textbf{break}
            }
            
        }
    }
    
    \Return{Updated $B$}
\end{algorithm}

\subsection{Local Search on the Problem Domain}

After the initial solution construction and post-processing, an item relocation local search is applied to further improve solution quality. This procedure iteratively attempts to move items between bins to reduce the total cost, considering both bin costs and link penalties. 
In each iteration, a subset of bins is selected with a given probability $p$, and all items within these bins are added to a candidate list. To prioritize moves most likely to yield a cost reduction, the candidates are sorted primarily by their aggregate neighbor weight in descending order, followed by the cost of their current bin (descending), and finally by the number of items in the bin (ascending).
For each candidate item, the algorithm evaluates all other opened bins to find a feasible destination and simulate a move. If it results in a lower total cost, the move is accepted and the solution is updated. This process continues until no further improvements are found or a maximum number of iterations is reached. After the relocation phase, the bin merging procedure is reapplied to further reduce costs. The Algorithm \ref{alg:item_relocation} describes the item relocation process.

\begin{algorithm}[!htb]
\footnotesize
\caption{Item Relocation Local Search}
\label{alg:item_relocation}
\KwIn{Set of opened bins $B$, probability $p$, max iterations $t_{max}$}

\For{$t = 1$ \KwTo $t_{max}$ \textbf{and} improved is True}{
    Set \texttt{improved} $\gets$ \texttt{false}; t++\;
    Initialize candidate list $\mathcal{C}$\;
    
    \For{each bin $j$ in $B$}{
        Draw $\rho \sim \mathrm{Uniform}(0,1)$\;
        \If{$\rho \leq p$}{
            \For{each item $i$ in $j$}{
                Add $(i, j)$ to $\mathcal{C}$\;
            }
        }
    }
    Sort $\mathcal{C}$ in descending order by $w_i^+$, then bin $c_k$, then ascending by number of items in $j$\;
    \For{each candidate $(i,j)$ in $\mathcal{C}$}{
        \For{each other bin $j'$ in $B$}{
            \If{$i$ fits in $j'$}{
                Simulate moving the item $i$ from bin $j$ to $j'$\;
                Apply bin type reduction to $j$ and $j'$\;
                $\Delta cost \gets$ simulated $B$ cost - current $B$ cost\;
                \If{$\Delta$cost $< 0$}{
                    Accept move: update $B$\;
                    Set \texttt{improved} $\gets$ \texttt{true}\;
                    \textbf{break} (first-improvement)\;
                }
            }
        }
        \If{\texttt{improved}}{\textbf{break}}
    }
    \If{not \texttt{improved}}{\textbf{break}}    
}
Apply bin merging (Algorithm~\ref{alg:bin_merging})\;
\Return{Updated $B$}
\end{algorithm}

\subsection{Objective Function Tie Breakers}

To guide the search toward more desirable solutions when objective function values are equal, three tie-breaking rules are applied in sequence. First, solutions that use bins with lower relative capacity utilization are preferred, promoting flexibility for accommodating future items. Second, among bins of equal utilization, those offering a better cost-to-capacity ratio are favored to reduce overall costs. 
Finally, solutions that resolve a greater link cost, thereby placing more highly connected items together, are prioritized, as this improves both feasibility and cost-effectiveness in the long term.

\section{Computational Experiments} \label{sec:results}

The set of 96 benchmark instances for the QMC-VSBPP used in this study were proposed by \cite{Meng2022}. Each instance defines a set of items with multidimensional weights, inter-item costs, and a collection of bin types with distinct capacities and fixed costs. The instances differ in the number of items ($n={25,50,100,200}$), bin types ($m={10,20,50}$), weight and capacity dimensionality ($d={3,5}$), and cost function (B1 - linear, B2 - concave, B3 - convex, and B4 - mixed). In this study, we adopt the same instances to ensure fair comparison and reproducibility.

The algorithms were implemented in C++ and executed on an Intel Core Ultra 9 185H 5.1 GHz processor with 64 GB of RAM.

Exact solutions for the QMC-VSBPP, using both the original and linearized models, were obtained with the Gurobi 12.0.2 solver \citep{gurobi}, with a time limit of 3600 seconds per instance. 

For the approximated approach using the RKO-ACO, each instance was run 30 times with the following configuration: best improvement local search, Q-learning for online parameter tuning, and 16 threads running the ACO metaheuristic. Time limits were set to 200 seconds for instances with $n=\{25,50\}$, 400 seconds for $n=100$, and 600 seconds for $n=200$, with a restart triggered at half of the maximum execution time. The solution pool size was set to 10.


ACO parameters were dynamically adjusted using Q-learning within the following discrete sets: $\kappa \in \{25,30\}$ for $n = \{25,50\}$ and $\kappa \in \{55,60\}$ for $n = \{100,200\}$, number of ants $\in \{2,3\}$, $q \in \{0.0001,0.001,0.1,0.3\}$, and $\xi \in \{0.80,0.85\}$.

To evaluate the quality of solutions, three commonly used metrics are employed. The Best Relative Percentage Deviation (BRPD) measures the deviation of the best solution obtained (Best) from the Best Known Solution (BKS) and is calculated as:

\begin{equation}
    \text{BRPD} = \frac{\text{Best} - \text{BKS}}{\text{Best}} \times 100\%,
\end{equation}

\noindent where the BKS is the minimum best solution reported either by Gurobi or in the literature.

The Average Relative Percentage Deviation (ARPD) captures the average deviation across $N$ independent runs and measures the performance consistency of the algorithm:

\begin{equation}
    \text{ARPD} = \frac{1}{N} \sum^N_{i=0} \frac{S_{i} - \text{BKS}}{S_{i}} \times 100\%,
\end{equation}

The Gap quantifies the difference between the best solution obtained and a lower bound (LB) found by the Gurobi using the proposed models, offering insight into the solution’s proximity to optimality:

\begin{equation}
    \text{Gap} = \frac{\text{Best} - \text{LB}}{\text{Best}} \times 100\%.
\end{equation}

Detailed instance-wise results are reported in the \ref{app_gurobi} and \ref{app_rko}, containing the instance settings, the best solution found by Gurobi and RKO-ACO, the time to the best solution, lower bound, Gap, ARPD, and BRPD values.

\subsection{Exact Results}

\begin{figure}[!htb]
    \centering
    \begin{subfigure}[t]{0.495\linewidth}
        \centering
        \includegraphics[width=\linewidth]{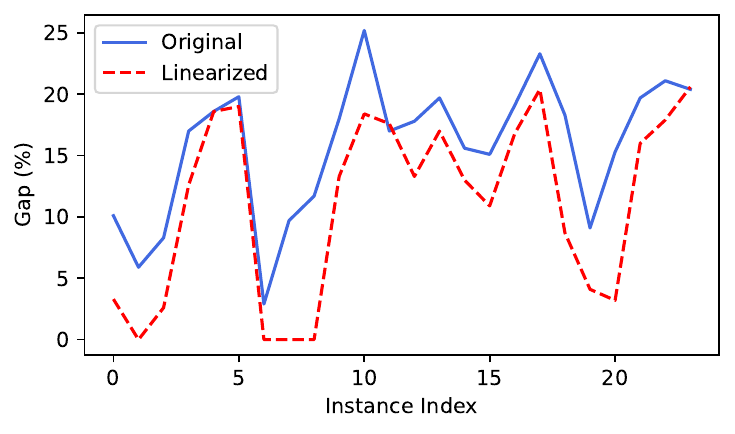}
        \caption{$n=25$}
        \label{fig:gurobi_gap_n25}
    \end{subfigure}
    \hfill
    \begin{subfigure}[t]{0.495\linewidth}
        \centering
        \includegraphics[width=\linewidth]{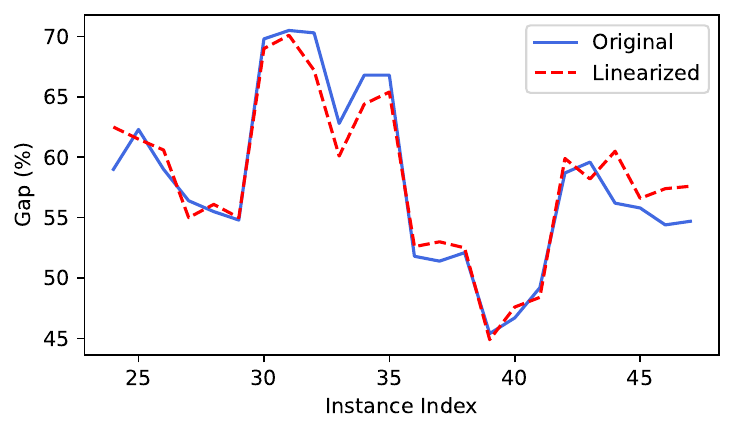}
        \caption{$n=50$}
        \label{fig:gurobi_gap_n50}
    \end{subfigure}

    \vspace{0.3cm}
    
    \begin{subfigure}[t]{0.495\linewidth}
        \centering
        \includegraphics[width=\linewidth]{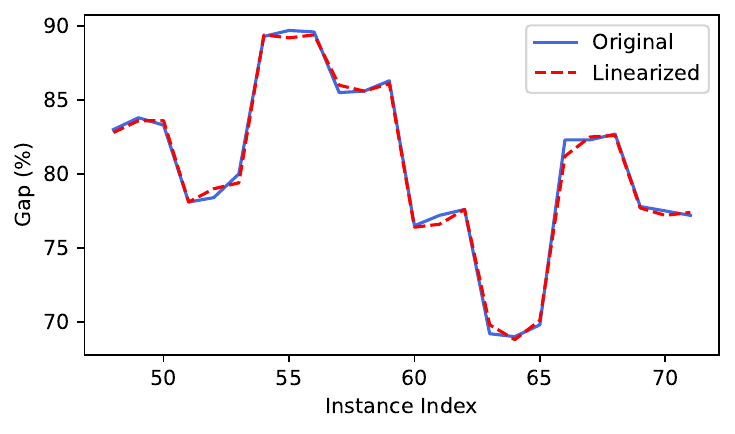}
        \caption{$n=100$}
        \label{fig:gurobi_gap_n100}
    \end{subfigure}
    \hfill
    \begin{subfigure}[t]{0.495\linewidth}
        \centering
        \includegraphics[width=\linewidth]{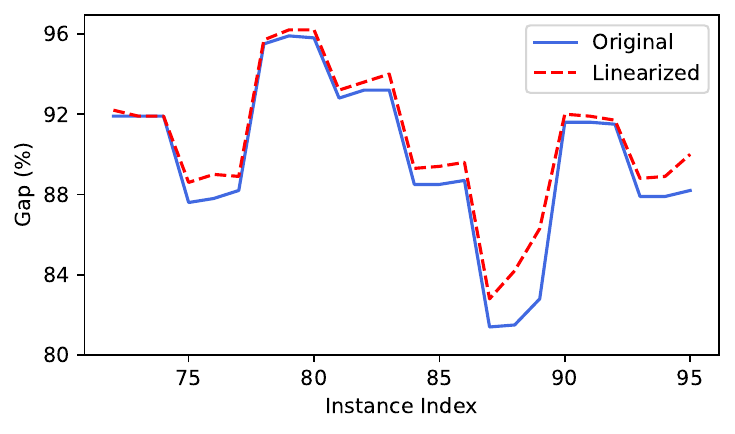}
        \caption{$n=200$}
        \label{fig:gurobi_gap_n200}
    \end{subfigure}
    
    \caption{Percentage gap from Gurobi results with original and linearized models}
    \label{fig:gurobi_gap}
\end{figure}

Both the original and linearized models of the QMC-VSBPP were solved using Gurobi. Figure~\ref{fig:gurobi_gap} shows the resulting gap values.

The results show that the original model is effective at finding high-quality solutions, particularly for larger instances, achieving superior solutions in 62\% of all cases. In contrast, the linearized model excels at providing tighter lower bounds and finding optimality in smaller instances (4 optimal solutions in $n=25$), outperforming the original model in 67\% of the instances in terms of lower bound quality. 

Nevertheless, for larger instances, both models exhibit substantial optimality gaps, averaging around 90\% with a one-hour execution limit, highlighting the computational difficulty of the QMC-VSBPP and motivating the adoption of advanced metaheuristics.

\subsection{RKO-ACO Results}

The RKO-ACO solutions were compared with the VNS results reported by \citet{Meng2022} and with the best solutions obtained by Gurobi for both the original and linearized models.

Table \ref{tab:result_summary} summarizes the computational performance of the evaluated solvers grouped by instance size ($n$). For each solver, the table reports the average best objective value found (Avg. Best), the average optimality gap (Avg. Gap(\%)), the number of instances where the method successfully found the best-known solution (\# BKS), and the average computational time in seconds (Avg. Time(s)). The best results are highlighted in bold.

\begin{table}[!htb]
\footnotesize
\caption{Performance comparison of Gurobi, VNS, and RKO-ACO models}
\begin{tabular}{llrrrr} \hline
 & Solvers & Avg. Best & Avg. Gap(\%) & \# BKS & Avg. Time(s) \\ \hline
\multirow{4}{*}{$n=25$} & Gurobi/orig. model & 9758 & 15.8 & 9 & 3600 \\
 & Gurobi/lin. model & 9799 & 11.1 & 11 & 3600 \\
 & VNS & 10192 & 15.3 & 0 & \textbf{1} \\
 & RKO-ACO & \textbf{9587} & \textbf{9.6} & \textbf{24} & 4 \\ \hline
\multirow{4}{*}{$n=50$} & Gurobi/orig. model & 29765 & 57.9 & 0 & 3600 \\
 & Gurobi/lin. model & 29998 & 58.2 & 0 & 3600 \\
 & VNS & 30671 & 58.4 & 0 & \textbf{2} \\
 & RKO-ACO & \textbf{28677} & \textbf{55.4} & \textbf{24} & 24 \\ \hline
\multirow{4}{*}{$n=100$} & Gurobi/orig. model & 113769 & 80.5 & 0 & 3600 \\
 & Gurobi/lin. model & 113785 & 80.4 & 0 & 3600 \\
 & VNS & 110026 & 79.7 & 0 & \textbf{16} \\
 & RKO-ACO & \textbf{103585} & \textbf{78.5} & \textbf{24} & 111 \\ \hline
\multirow{4}{*}{$n=200$} & Gurobi/orig. model & 395891 & 89.8 & 1 & 3600 \\
 & Gurobi/lin. model & 430017 & 90.7 & 0 & 3600 \\
 & VNS & 410628 & 90.2 & 0 & \textbf{109} \\
 & RKO-ACO & \textbf{382978} & \textbf{89.5} & \textbf{23} & 231 \\ \hline
\multirow{4}{*}{All} & Gurobi/orig. model & 137296 & 61.0 & 10 & 3600 \\
 & Gurobi/lin. model & 145900 & 60.1 & 11 & 3600 \\
 & VNS & 140379 & 60.9 & 0 & \textbf{32} \\
 & RKO-ACO & \textbf{131207} & \textbf{58.2} & \textbf{95} & 92 \\ \hline
\end{tabular}
\label{tab:result_summary}
\end{table}

According to Table~\ref{tab:result_summary}, VNS shows the weakest performance in terms of solution quality. This method failed to achieve the BKS in any of the evaluated instances (0 out of 96), presenting average optimality gaps that range from 15.3\% for $n=25$ up to 90.2\% for $n=200$, although it requires the least computational time.

Furthermore, the Gurobi solver with the original and linearized models demonstrates severe limitations as instance complexity grows. While they achieve the BKS in nearly half of the $n=25$ instances (9 and 11, respectively), with the linearized model successfully finding 4 proven optimal solutions, they completely fail to find the best solutions for $n=50$ and $n=100$ before reaching the 3600-second time limit. In the largest case ($n=200$), the original model finds the BKS in only a single instance.

The proposed RKO-ACO demonstrates strong performance across all instance sizes, consistently outperforming the baselines. For $n=25$, it attains the BKS in all 24 instances; notably, it matches the 4 exact optimal solutions identified by Gurobi, while establishing better upper bounds for the remaining cases. For $n=50$ and $n=100$, it achieves the BKS in all 48 instances, surpassing both exact and heuristic methods from the literature. In the largest case, $n=200$, it achieves the BKS in 23 out of 24 instances, further confirming its robustness and reaching an overall success rate of 95 out of 96 instances in a fraction of the time required by the exact solver.

\begin{figure}[!htb]
    \centering
    \begin{subfigure}[t]{0.495\linewidth}
        \centering
        \includegraphics[width=\linewidth]{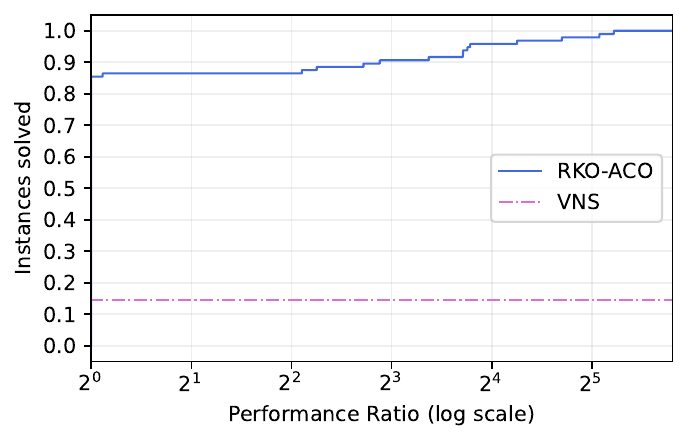}
        \caption{Performance Profile 5\%}
        \label{fig:perf_profile_5}
    \end{subfigure}
    \hfill
    \begin{subfigure}[t]{0.495\linewidth}
        \centering
        \includegraphics[width=\linewidth]{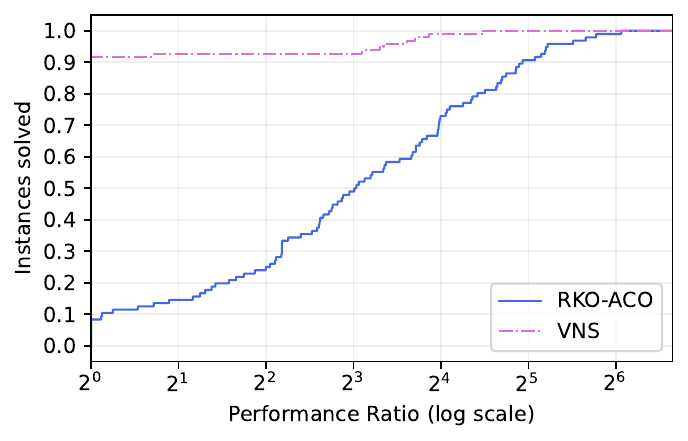}
        \caption{Performance Profile without target solution}
        \label{fig:perf_profile_100}
    \end{subfigure}
    \caption{Performance profiles for different tolerance levels}
    \label{fig:perf_profile}
\end{figure}

Figure \ref{fig:perf_profile} shows the performance profile charts comparing the VNS and RKO-ACO methods in terms of computational time.

Figure~\ref{fig:perf_profile_5} presents the performance profile comparing the time required by VNS and RKO-ACO to reach a solution within 5\% of the best-known value. To ensure a fair comparison, the VNS runtimes were adjusted by a factor of 2.2 to compensate for hardware performance differences, according to PassMark benchmarks\footnote{\url{https://www.cpubenchmark.net/}}. The results reveal a significant disparity in both efficiency and robustness. At the baseline ratio ($2^0$), RKO-ACO was the fastest algorithm to reach the target solution in 85\% of instances, while VNS was faster in 15\%. Furthermore, the profile shows that while RKO-ACO ultimately achieves the target solution in 100\% of instances as the time ratio increases, VNS performance plateaus, failing to improve beyond its initial 15\% success rate, regardless of the increased time allowance. The performance profile shown in Figure~\ref{fig:perf_profile_100} disregards the target solution quality. In this scenario, VNS outperforms RKO-ACO in computational time.

\begin{figure}[!htb]
    \centering
    \begin{subfigure}[t]{0.495\linewidth}
        \centering
        \includegraphics[width=\linewidth]{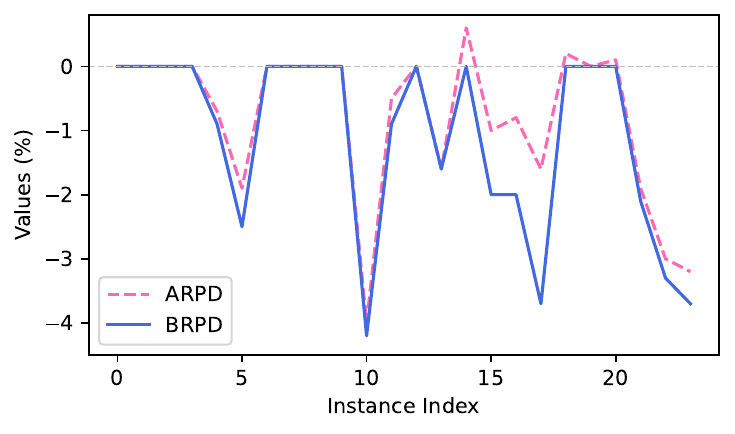}
        \caption{$n=25$}
        \label{fig:rko_arpd_n25}
    \end{subfigure}
    \hfill
    \begin{subfigure}[t]{0.495\linewidth}
        \centering
        \includegraphics[width=\linewidth]{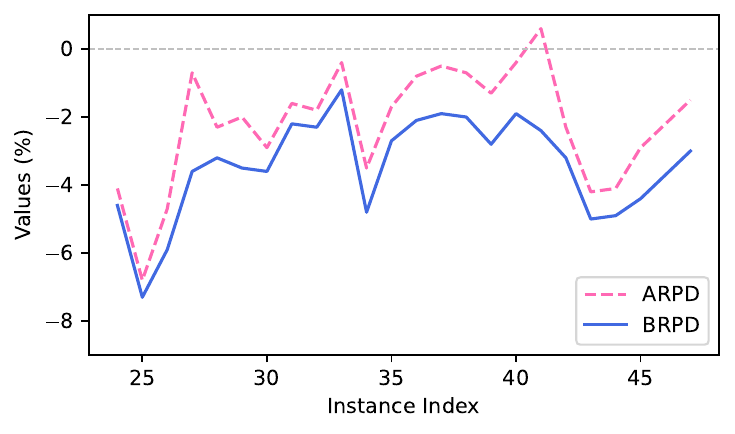}
        \caption{$n=50$}
        \label{fig:rko_arpd_n50}
    \end{subfigure}

    \vspace{0.3cm}
    
    \begin{subfigure}[t]{0.495\linewidth}
        \centering
        \includegraphics[width=\linewidth]{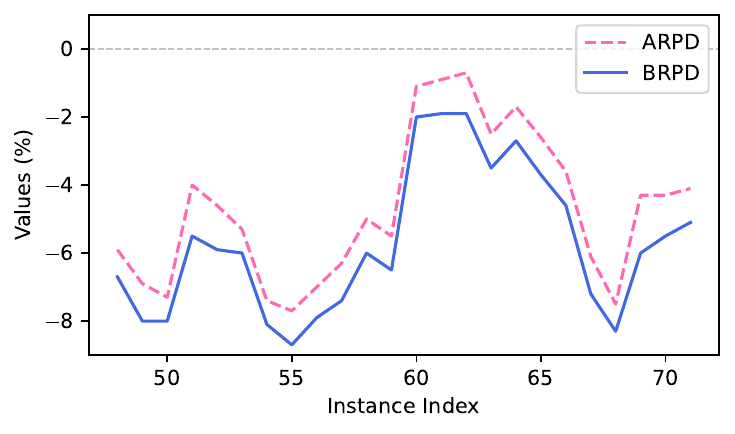}
        \caption{$n=100$}
        \label{fig:rko_arpd_n100}
    \end{subfigure}
    \hfill
    \begin{subfigure}[t]{0.495\linewidth}
        \centering
        \includegraphics[width=\linewidth]{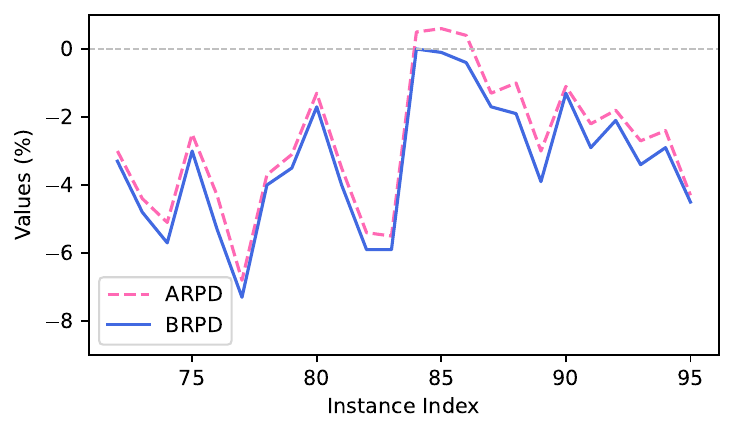}
        \caption{$n=200$}
        \label{fig:rko_arpd_n200}
    \end{subfigure}
    
    \caption{RKO-ACO's ARPD and BRPD values}
    \label{fig:rko_arpd}
\end{figure}

In Figure~\ref{fig:rko_arpd}, the RKO-ACO's ARPD and BRPD values are reported for each instance set. A larger difference between ARPD and BRPD reflects greater variance across runs, while negative values indicate improvements over the BKS.

As shown in Figure~\ref{fig:rko_arpd}, the average difference between ARPD and BRPD is 0.8\% with a standard deviation of 0.6\%. The largest differences occur for instances with $n=50$, where the ARPD exceeds the BRPD by up to 3\% and 1.2\% on average. In contrast, for instances with $n=200$, the maximum difference is 1\% and the average is 0.5\%. Overall, the ARPD remains below or equal to zero in 93\% of the instances, and the BRPD remains below or equal to zero in 100\% of the cases, confirming that the algorithm consistently matches or improves upon the BKS.

\begin{figure}[!htb]
    \centering
    \includegraphics[width=0.6\linewidth]{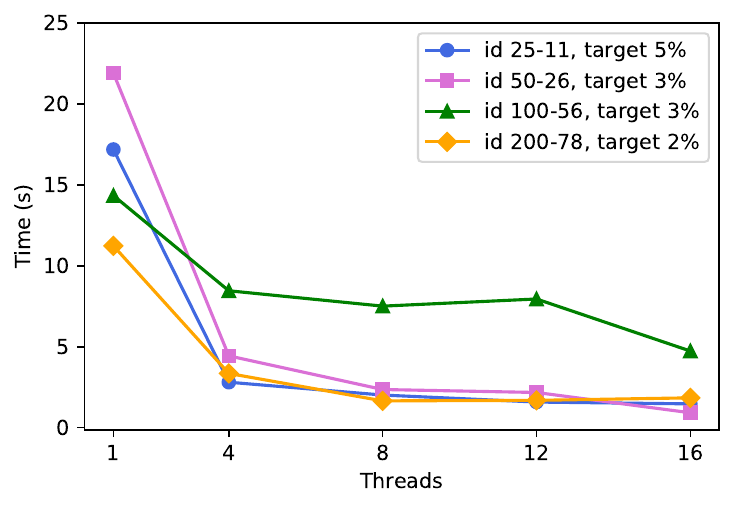}
    \caption{RKO-ACO's time to target solution by number of threads}
    \label{fig:ttt_threads}
\end{figure}

The time analysis of RKO-ACO per thread configuration, where each thread runs an independent instance of the ACO algorithm, is presented in Figure~\ref{fig:ttt_threads}. For each value of $n$, the best-performing instance was selected to measure the computational time (in seconds) required to reach solution targets within 2\% to 5\% of the best RKO-ACO result for that instance. With a single thread, the time required exceeds 11 s. Using four threads leads to a substantial reduction, with an average time of 5 s. With eight threads, the target solutions are obtained in 7.5 s for instances with $n=50$, and approximately 2 s for the remaining instances.

\begin{figure}[!htb]
    \centering
    \includegraphics[width=1\linewidth]{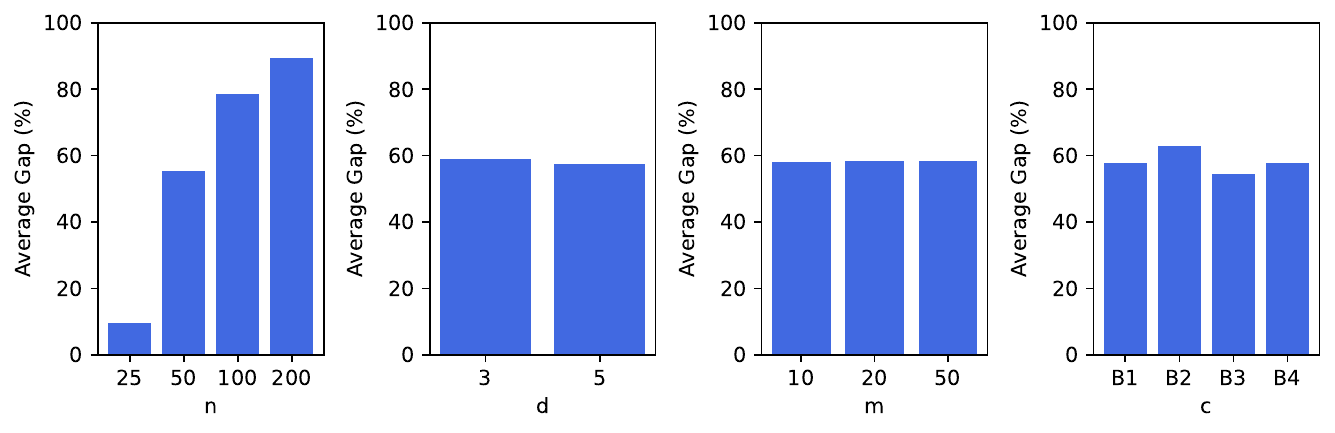}
    \caption{Average gap from RKO-ACO by instance settings: $n$, $d$, $m$, and $c$}
    \label{fig:rko_gap_params}
\end{figure}

In addition, we evaluate the RKO-ACO’s ability to find high-quality solutions by verifying the average gap per instance setting: $n$ (number of items), $m$ (number of bins), $d$ (weight and capacity dimensions), and $c$ (cost function: B1 - linear, B2 - concave, B3 - convex, or B4 - mixed), as presented in Figure~\ref{fig:rko_gap_params}.

As shown in Figure~\ref{fig:rko_gap_params}, the main factor contributing to the solution gap is the number of items, indicating that the gap is directly proportional to $n$. This is supported by a Pearson correlation of 0.83 between gap and $n$ values. Although the cost function B2 (concave) appears to impact the gap, an ANOVA test indicates that this difference is not statistically significant (p-value $ = 0.83$).

\subsection{Statistical Analysis}

Since the Shapiro–Wilk test indicated non-normality for all paired differences (p-value $< 10^{-5}$), nonparametric tests were applied. The Friedman test detected significant performance differences among the methods (p-value $= 5.4 \times 10^{-34}$). Pairwise Wilcoxon tests confirmed that RKO-ACO significantly outperforms both VNS (p-value $= 8.8 \times 10^{-18}$) and Gurobi (p-value $= 2.7 \times 10^{-15}$).

\subsection{Ablation Study}

To isolate the individual contributions of RKO-ACO’s components, an ablation study was conducted to compare the full algorithm against three variants: RKO-ACO without Q-learning, RKO-ACO without Nelder-Mead local search, and the standalone continuous-domain ACO metaheuristic. 

The study focused on three of the best-performing large-scale instances ($n=200$), with results averaged over 30 independent runs. Parameters were tuned using \texttt{irace} \citep{irace}, resulting in the following configurations: two ants per generation, $\kappa = 55$, $q = 0.0001$, and $\xi = 0.8$. These parameters were applied for both the variant without Q-learning and the  Continuous ACO.
The RKO-ACO versions utilized parallelization with 16 threads, and the Continuous ACO was executed in a single thread. 
The results are summarized in Table~\ref{tab:ablation}, detailing the best solution found (Best), the percentage difference (Diff) of each variant relative to the best solution found by the full RKO-ACO, and the Wilcoxon paired test p-values.

\begin{table}[htb]
\centering
\footnotesize
\caption{Ablation study}
\begin{tabular}{l|ll|ll|ll|l}
\hline
Instance & \multicolumn{2}{c|}{200-78} & \multicolumn{2}{c|}{200-83} & \multicolumn{2}{c|}{200-84} & \multirow{2}{*}{p-value} \\ \cline{1-7}
Version & Best & Diff & Best & Diff & Best & Diff &  \\ \hline
Continuous ACO & 403052 & 2.6\% & 373832 & 2.2\% & 370767 & 2.1\% & 0.0014 \\
w/o Nelder-Mead & 398065 & 1.3\% & 369726 & 1.1\% & 367290 & 1.1\% & 0.0015 \\
w/o Q-Learning & 395053 & 0.6\% & 367134 & 0.4\% & 364773 & 0.4\% & $0.56$ \\ 
\textbf{RKO-ACO} & \textbf{392877} & - & \textbf{365656} & - & \textbf{363227} & - &  - \\ \hline
\end{tabular}
\label{tab:ablation}
\end{table}

The ablation study demonstrated that the full RKO-ACO algorithm consistently outperforms all variants across the tested instances. The Continuous ACO produced the lowest solution quality, with a mean degradation of 2.3\% compared to the full version. Removing the Nelder-Mead local search resulted in a performance drop of approximately 1.15\%, highlighting its essential role in refining candidate solutions within the continuous space. Lastly, the version without Q-learning yields solutions that are 0.5\% worse on average than the full RKO-ACO. 

Using Friedman test (p-value $= 7.4 \times 10^{-24}$) followed by Wilcoxon signed-rank tests, we have statistically validated that the full RKO-ACO significantly outperforms the standalone Continuous ACO (p-value $= 1.4 \times 10^{-3}$) and the version without Nelder-Mead local search (p-value $= 1.5 \times 10^{-3}$). While the Q-learning component contributes to finding the best individual solutions, the statistical difference was not significant (p-value $=0.56$).

\section{Conclusion} \label{sec:conclusion}

This study advances the state-of-the-art for the Quadratic Multiple Constraints Variable-Sized Bin Packing Problem (QMC-VSBPP) by integrating exact and metaheuristic solution approaches.

First, we proposed a linearization of the original model, which contained quadratic terms. This linearized model facilitated the use of exact solvers and produced tighter lower bounds, providing the first reported lower-bound results for the QMC-VSBPP. Although optimality gaps remain significant for larger instances, the linearized model establishes a stronger foundation for exact optimization.

The proposed RKO-ACO metaheuristic demonstrated robust performance across all benchmark instances, consistently achieving or surpassing best-known solutions from both exact and heuristic methods from the literature. Notably, the RKO-ACO improved 95 of 96 benchmark instances, confirming its effectiveness in handling the combinatorial complexity of the problem. Analysis of ARPD and BRPD metrics further validated the stability of the approach across independent runs.

The results of this study confirm the research hypotheses introduced in Section~\ref{sec:intro}, as detailed below.

\textbf{(1) Effectiveness of linearization.} The linearized model of the QMC-VSBPP consistently resulted in tighter lower bounds compared to the original quadratic model. Also, this simplified model could obtain four optimal solutions in small instances. However, both models exhibited substantial optimality gaps for larger instances, reflecting the intrinsic combinatorial complexity of the problem and motivating the need for complementary metaheuristic strategies.

\textbf{(2) Performance of RKO-ACO.} The RKO-ACO algorithm demonstrated robust performance across all tested instances, outperforming best-known solutions from both exact and traditional methods from the literature. Analysis of ARPD and BRPD metrics across independent runs confirmed the reliability and stability of the approach.

\textbf{(3) Comparison with benchmarks.} The exact Gurobi solutions already outperform the VNS results reported by \citet{Meng2022} in most of the instances. However, the proposed RKO-ACO further improves upon both, achieving the best-known solution or establishing new upper bounds for all $n=25$ to $n=100$ cases, and 92\% of instances with $n=200$. These results establish updated benchmark values for future studies.

Overall, the combination of exact and metaheuristic methods provides complementary benefits: the model linearization improves lower bound estimation and simplifies exact optimization approaches, while RKO-ACO effectively explores the solution space, obtaining high-quality solutions with computational efficiency. These findings demonstrate that adaptive metaheuristics, particularly when integrated with continuous-domain representations, are highly effective for solving complex quadratic packing problems.

Future work may explore hybrid exact–heuristic methods and exploit problem-specific structures to further enhance scalability.
Additionally, future research will focus on the framework scalability for large-scale datasets using GPU-based architectures and batch processing.

\section*{Author contributions: CRediT}

\textbf{Natalia Alves Santos:} Conceptualization, Methodology, Software, Visualization, Writing – Original Draft. \textbf{Marlon Jeske:} Methodology, Validation, Writing – Review and Editing, Funding acquisition. \textbf{Antônio Augusto Chaves:} Conceptualization, Supervision, Resources, Software, Validation, Writing – Review and Editing, Funding acquisition.

\section*{Funding sources}

This study was financed in part by the Brazilian Federal Agency for Support and Evaluation of Graduate Education (CAPES) - Finance Code 001.
Marlon Jeske was supported by the S\~ao Paulo Research Foundation (FAPESP) under grant 2025/00386-3.
Antonio Chaves was supported by the S\~ao Paulo Research Foundation (FAPESP) under grants 2022/05803-3 and 2024/08848-3, and the National Council for Scientific and Technological Development (CNPq) under grant 305557/2024-68.



\section*{Data and Code Availability}

The RKO-ACO source code and instances to reproduce the experiments are publicly available at \url{https://github.com/nataliaalves03/RKO-ACO}.



\appendix

\section{Gurobi results from original and linearized models}
\label{app_gurobi}

{\footnotesize
\begin{longtable}{ll|rrr|rrr}
\caption{Gurobi results} \label{tab:gurobi} \\
\hline
    \multicolumn{2}{c}{Instance} & \multicolumn{3}{|c|}{Original model} & \multicolumn{3}{c}{Linearized model} \\ \hline
   ID &   $n$ &   LB &   Best &   Gap(\%) &   LB &   Best &   Gap(\%) \\
\hline
 \endfirsthead
\hline
    \multicolumn{2}{c}{Instance} & \multicolumn{3}{|c|}{Original model} & \multicolumn{3}{c}{Linearized model} \\ \hline
   ID &   $n$ &   LB &   Best &   Gap(\%) &   LB &   Best &   Gap(\%) \\
\hline
\endhead
\hline
\endfoot
\endlastfoot

 1    & 25  & 6519            & \textbf{7253}     & 10.1             & \textbf{7014}  & \textbf{7253}            & \textbf{3.3}   \\
 2    & 25  & 6557            & \textbf{6968}     & 5.9              & \textbf{6968}  & \textbf{6968}            & \textbf{0.0}   \\
 3    & 25  & 6438            & 7022              & 8.3              & \textbf{6758}  & \textbf{6935}   & \textbf{2.6}   \\
 4    & 25  & 9437            & \textbf{11371}    & 17.0             & \textbf{9979}  & 11415           & \textbf{12.6}  \\
 5    & 25  & 9023            & \textbf{11082}    & \textbf{18.6}    & \textbf{9199}  & 11304           & \textbf{18.6}  \\
 6    & 25  & 8919            & \textbf{11123}    & 19.8    & \textbf{9087}  & 11211           & \textbf{19.0}  \\
 7    & 25  & 4976            & \textbf{5125}     & 2.9              & \textbf{5125}  & \textbf{5125}            & \textbf{0.0}   \\
 8    & 25  & 4484            & \textbf{4967}     & 9.7              & \textbf{4967}  & \textbf{4967}            & \textbf{0.0}   \\
 9    & 25  & 4385            & \textbf{4965}     & 11.7             & \textbf{4965}  & \textbf{4965}            & \textbf{0.0}   \\
 10   & 25  & 6322            & \textbf{7713}     & 18.0             & \textbf{6685}  & \textbf{7713}            & \textbf{13.3}  \\
 11   & 25  & 5689            & \textbf{7603}     & 25.2             & \textbf{6262}  & 7679            & \textbf{18.4}  \\
 12   & 25  & 5960            & \textbf{7176}     & \textbf{17.0}    & \textbf{6134}  & 7441            & 17.6  \\
 13   & 25  & 8868            & \textbf{10787}    & 17.8             & \textbf{9353}  & \textbf{10787}           & \textbf{13.3}  \\
 14   & 25  & 8390            & \textbf{10443}    & 19.7             & \textbf{8881}  & 10695           & \textbf{17.0}  \\
 15   & 25  & 8597            & \textbf{10185}    & 15.6             & \textbf{9101}  & 10463           & \textbf{13.0}  \\
 16   & 25  & 15276           & \textbf{17990}    & 15.1             & \textbf{16057} & 18021           & \textbf{10.9}  \\
 17   & 25  & 14364           & \textbf{17756}    & 19.1             & \textbf{14898} & 17897           & \textbf{16.8}  \\
 18   & 25  & 13268           & 17308             & 23.3             & \textbf{13608} & \textbf{17093}  & \textbf{20.4}  \\
 19   & 25  & 6674            & 8171              & 18.3             & \textbf{7258}  & \textbf{7949}   & \textbf{8.7}   \\
 20   & 25  & 7041            & 7748              & 9.1              & \textbf{7376}  & \textbf{7689}   & \textbf{4.1}   \\
 21   & 25  & 6608            & 7803              & 15.3             & \textbf{7370}  & \textbf{7615}   & \textbf{3.2}   \\
 22   & 25  & 9123            & \textbf{11361}    & 19.7             & \textbf{9678}  & 11521           & \textbf{16.0}  \\
 23   & 25  & 8896            & 11274             & 21.1             & \textbf{9139}  & \textbf{11130}  & \textbf{17.9}  \\
 24   & 25  & 8764            & \textbf{11003}    & \textbf{20.4}    & \textbf{8999}  & 11330           & 20.6  \\
 \hline

 25   & 50  & \textbf{10359}  & \textbf{25237}    & \textbf{59.0}    & 9595           & 25616           & 62.5           \\
 26   & 50  & \textbf{9406}   & 24922             & 62.3             & \textbf{9406}           & \textbf{24455}  & \textbf{61.5}  \\
 27   & 50  & \textbf{9950}   & 24259             & \textbf{59.0}    & 9434           & \textbf{23951}  & 60.6           \\
 28   & 50  & 14451           & 33174             & 56.4             & \textbf{14607} & \textbf{32491}  & \textbf{55.0}  \\
 29   & 50  & \textbf{14661}  & \textbf{32915}    & \textbf{55.5}    & 14476          & 33000           & 56.1           \\
 30   & 50  & 14568           & \textbf{32203}    & \textbf{54.8}    & \textbf{14791} & 32863           & 55.0           \\
 31   & 50  & 6344            & 20988             & 69.8    & \textbf{6488}  & \textbf{20966}  & \textbf{69.0}  \\
 32   & 50  & 5566            & \textbf{18854}    & 70.5    & \textbf{5821}  & 19493           & \textbf{70.1}  \\
 33   & 50  & 5608            & \textbf{18876}    & 70.3             & \textbf{6423}  & 19585           & \textbf{67.2}  \\
 34   & 50  & 9440            & \textbf{25377}    & 62.8             & \textbf{10192} & 25535           & \textbf{60.1}  \\
 35   & 50  & 8614            & \textbf{25977}    & 66.8             & \textbf{9250}  & 25984           & \textbf{64.4}  \\
 36   & 50  & 8218            & \textbf{24771}    & 66.8             & \textbf{8858}  & 25571           & \textbf{65.4}  \\
 37   & 50  & \textbf{15323}  & 31790             & \textbf{51.8}    & 14942          & \textbf{31508}  & 52.6           \\
 38   & 50  & \textbf{15323}  & \textbf{31505}    & \textbf{51.4}    & 14853          & 31603           & 53.0           \\
 39   & 50  & \textbf{14984}  & \textbf{31299}    & \textbf{52.1}    & 14902          & 31355           & 52.5  \\
 40   & 50  & 24973           & 45695             & 45.4             & \textbf{25085} & \textbf{45507}  & \textbf{44.9}  \\
 41   & 50  & \textbf{24434}  & 45819             & \textbf{46.7}    & 23923          & \textbf{45670}  & 47.6           \\
 42   & 50  & \textbf{22912}  & 45103             & 49.2             & 22617          & \textbf{43860}  & \textbf{48.4}  \\
 43   & 50  & \textbf{10814}  & \textbf{26210}    & \textbf{58.7}    & 10719          & 26703           & 59.9           \\
 44   & 50  & 10426           & \textbf{25813}    & 59.6             & \textbf{10919} & 26148           & \textbf{58.2}  \\
 45   & 50  & \textbf{11238}  & \textbf{25647}    & \textbf{56.2}    & 10375          & 26294           & 60.5           \\
 46   & 50  & 14475           & \textbf{32768}    & \textbf{55.8}    & \textbf{15175} & 34930           & 56.6           \\
 47   & 50  & \textbf{14973}  & \textbf{32833}    & \textbf{54.4}    & 14161          & 33205           & 57.4           \\
 48   & 50  & \textbf{14630}  & \textbf{32316}    & \textbf{54.7}    & 14267          & 33651           & 57.6           \\
\hline

 49   & 100 & 17153           & 100850            & 83.0             & \textbf{17217} & \textbf{100173} & \textbf{82.8}  \\
 50   & 100 & \textbf{17120}  & 105832            & 83.8    & 17115          & \textbf{104678} & \textbf{83.6}  \\
 51   & 100 & \textbf{17111}  & \textbf{102547}   & \textbf{83.3}    & 17110          & 104330          & 83.6  \\
 52   & 100 & 27223           & \textbf{124056}   & \textbf{78.1}    & \textbf{27533} & 125941          & \textbf{78.1}  \\
 53   & 100 & \textbf{27237}  & \textbf{126271}   & \textbf{78.4}    & 27190          & 129477          & 79.0           \\
 54   & 100 & \textbf{27189}  & 136308            & 80.0             & 27185          & \textbf{131648} & \textbf{79.4}  \\
 55   & 100 & \textbf{10218}  & 95354             & \textbf{89.3}    & 10154          & \textbf{95298}  & 89.4  \\
 56   & 100 & 9185            & 89120             & 89.7    & \textbf{9571}  & \textbf{88579}  & \textbf{89.2}  \\
 57   & 100 & 9271            & 89392             & 89.6    & \textbf{9296}  & \textbf{87901}  & \textbf{89.4}  \\
 58   & 100 & \textbf{16074}  & \textbf{111149}   & \textbf{85.5}    & 15755          & 112180          & 86.0           \\
 59   & 100 & 15577           & 108556            & \textbf{85.6}    & \textbf{15631} & \textbf{108368} & \textbf{85.6}  \\
 60   & 100 & 14859           & 108521            & 86.3    & \textbf{15104} & \textbf{108445} & \textbf{86.1}  \\
 61   & 100 & \textbf{26205}  & 111705            & 76.5    & 26169          & \textbf{111116} & \textbf{76.4}  \\
 62   & 100 & 25400           & 111573            & 77.2             & \textbf{26135} & \textbf{111494} & \textbf{76.6}  \\
 63   & 100 & 24847           & \textbf{110730}   & \textbf{77.6}    & \textbf{25066} & 112157          & \textbf{77.6}  \\
 64   & 100 & \textbf{46806}  & 152214            & \textbf{69.2}    & 44932          & \textbf{148884} & 69.8  \\
 65   & 100 & 45547           & 146962            & 69.0             & \textbf{45595} & \textbf{146146} & \textbf{68.8}  \\
 66   & 100 & 42899           & \textbf{142256}   & \textbf{69.8}    & \textbf{43378} & 144898          & 70.1           \\
 67   & 100 & 17880           & 101135            & 82.3             & \textbf{18905} & \textbf{100798} & \textbf{81.2}  \\
 68   & 100 & 17860           & \textbf{100817}   & \textbf{82.3}    & \textbf{18121} & 103431          & 82.5  \\
 69   & 100 & 17707           & \textbf{102307}   & 82.7    & \textbf{17810} & 102549          & \textbf{82.6}  \\
 70   & 100 & \textbf{26710}  & 120139            & 77.8    & 26541          & \textbf{118839} & \textbf{77.7}  \\
 71   & 100 & 26461           & 117592            & 77.5    & \textbf{26570} & \textbf{116803} & \textbf{77.2}  \\
 72   & 100 & 26268           & \textbf{115080}   & \textbf{77.2}    & \textbf{26330} & 116717          & 77.4  \\ 
\hline

 73   & 200 & \textbf{30470}  & \textbf{375104}   & \textbf{91.9}    & \textbf{30470}          & 393151          & 92.2           \\
 74   & 200 & \textbf{30470}  & \textbf{374153}   & \textbf{91.9}    & \textbf{30470}          & 374507          & \textbf{91.9}  \\
 75   & 200 & \textbf{30470}  & \textbf{375918}   & \textbf{91.9}    & \textbf{30470}          & 376089          & \textbf{91.9}  \\
 76   & 200 & \textbf{50565}  & \textbf{409442}   & \textbf{87.6}    & \textbf{50565}          & 443658          & 88.6           \\
 77   & 200 & \textbf{50565}  & \textbf{415919}   & \textbf{87.8}    & \textbf{50565}          & 457660          & 89.0           \\
 78   & 200 & \textbf{50565}  & \textbf{429660}   & \textbf{88.2}    & \textbf{50565}          & 456602          & 88.9  \\
 79   & 200 & \textbf{16107}  & \textbf{360694}   & \textbf{95.5}    & \textbf{16107}          & 375502          & 95.7  \\
 80   & 200 & 14459           & \textbf{350099}   & \textbf{95.9}    & \textbf{14460} & 376069          & 96.2           \\
 81   & 200 & \textbf{14442}  & \textbf{344132}   & \textbf{95.8}    & \textbf{14442}          & 377639          & 96.2           \\
 82   & 200 & \textbf{27553}  & \textbf{382758}   & \textbf{92.8}    & \textbf{27553}          & 405994          & 93.2           \\
 83   & 200 & \textbf{26310}  & \textbf{387238}   & \textbf{93.2}    & \textbf{26310}          & 414551          & 93.6  \\
 84   & 200 & \textbf{25987}  & \textbf{384555}   & \textbf{93.2}    & \textbf{25987}          & 435637          & 94.0           \\
 85   & 200 & \textbf{45272}  & \textbf{393751}   & \textbf{88.5}    & \textbf{45272}          & 424766          & 89.3           \\
 86   & 200 & \textbf{45229}  & \textbf{394673}   & \textbf{88.5}    & \textbf{45229}          & 428353          & 89.4           \\
 87   & 200 & \textbf{44800}  & \textbf{394898}   & \textbf{88.7}    & \textbf{44800}          & 429283          & 89.6           \\
 88   & 200 & 85027           & \textbf{457451}   & \textbf{81.4}    & \textbf{85028} & 495363          & 82.8           \\
 89   & 200 & 85027           & \textbf{459525}   & \textbf{81.5}    & \textbf{85028} & 539772          & 84.2           \\
 90   & 200 & \textbf{79786}  & \textbf{464271}   & \textbf{82.8}    & \textbf{79786}          & 581736          & 86.3           \\
 91   & 200 & 31556           & \textbf{373626}   & \textbf{91.6}    & \textbf{31557} & 395750          & 92.0           \\
 92   & 200 & 31556           & \textbf{374781}   & \textbf{91.6}    & \textbf{31557} & 391674          & 91.9  \\
 93   & 200 & 31490           & \textbf{371865}   & \textbf{91.5}    & \textbf{31491} & 379347          & 91.7  \\
 94   & 200 & 49357           & \textbf{408022}   & \textbf{87.9}    & \textbf{49358} & 438879          & 88.8           \\
 95   & 200 & \textbf{49196}  & \textbf{406270}   & \textbf{87.9}    & \textbf{49196}          & 444231          & 88.9           \\
 96   & 200 & 48657           & \textbf{412571}   & \textbf{88.2}    & \textbf{48658} & 484192          & 90.0           \\
\hline

\end{longtable}}

\section{RKO-ACO results}
\label{app_rko}

{\setlength{\tabcolsep}{5pt}
{\footnotesize
\begin{longtable}{lllll|rr|rrrrr}
\caption{RKO-ACO results}  \label{tab:rko_results} \\
\hline
 ID   &   $n$   & $c$   & $d$   & $m$   & LB    & BKS & RKO-ACO & Time(s)   & Gap  & BRPD   & ARPD   \\ \hline
 \endfirsthead
\hline
 ID   &   $n$   & $c$   & $d$   & $m$   & LB    & BKS  & RKO-ACO & Time(s)   & Gap  & BRPD   & ARPD    \\ \hline
\endhead
\hline
\endfoot
\endlastfoot
 1    & 25  & B1  & 3   & 10  & 7014  & \textbf{7253}   & \textbf{7253}   & 0.0         & 3.3       & 0.0    & 0.0    \\
 2    & 25  & B1  & 3   & 20  & 6968  & \textbf{6968}   & \textbf{6968}   & 0.0         & 0.0       & 0.0    & 0.0    \\
 3    & 25  & B1  & 3   & 50  & 6758  & \textbf{6935}   & \textbf{6935}   & 0.5         & 2.6       & 0.0    & 0.0    \\
 4    & 25  & B1  & 5   & 10  & 9979  & \textbf{11371}  & \textbf{11371}  & 15.1        & 12.2      & 0.0    & 0.0    \\
 5    & 25  & B1  & 5   & 20  & 9199  & 11082           & \textbf{10978}  & 26.7        & 16.2      & -0.9   & -0.7   \\
 6    & 25  & B1  & 5   & 50  & 9087  & 11123           & \textbf{10855}  & 23.0        & 16.3      & -2.5   & -1.9   \\
 7    & 25  & B2  & 3   & 10  & 5125  & \textbf{5125}   & \textbf{5125}   & 0.0         & 0.0       & 0.0    & 0.0    \\
 8    & 25  & B2  & 3   & 20  & 4967  & \textbf{4967}   & \textbf{4967}   & 0.0         & 0.0       & 0.0    & 0.0    \\
 9    & 25  & B2  & 3   & 50  & 4965  & \textbf{4965}   & \textbf{4965}   & 0.0         & 0.0       & 0.0    & 0.0    \\
 10   & 25  & B2  & 5   & 10  & 6685  & \textbf{7713}   & \textbf{7713}   & 0.2         & 13.3      & 0.0    & 0.0    \\
 11   & 25  & B2  & 5   & 20  & 6262  & 7603            & \textbf{7299}   & 3.3         & 14.2      & -4.2   & -4.0   \\
 12   & 25  & B2  & 5   & 50  & 6134  & 7176            & \textbf{7115}   & 0.1         & 13.8      & -0.9   & -0.5   \\
 13   & 25  & B3  & 3   & 10  & 9353  & \textbf{10787}  & \textbf{10787}  & 0.7         & 13.3      & 0.0    & 0.0    \\
 14   & 25  & B3  & 3   & 20  & 8881  & 10443           & \textbf{10276}  & 0.6         & 13.6      & -1.6   & -1.6   \\
 15   & 25  & B3  & 3   & 50  & 9101  & \textbf{10185}  & \textbf{10185}  & 0.8         & 10.6      & 0.0    & 0.6    \\
 16   & 25  & B3  & 5   & 10  & 16057 & 17990           & \textbf{17630}  & 4.2         & 8.9       & -2.0   & -1.0   \\
 17   & 25  & B3  & 5   & 20  & 14898 & 17756           & \textbf{17401}  & 0.4         & 14.4      & -2.0   & -0.8   \\
 18   & 25  & B3  & 5   & 50  & 13608 & 17093           & \textbf{16489}  & 2.6         & 17.5      & -3.7   & -1.6   \\
 19   & 25  & B4  & 3   & 10  & 7258  & \textbf{7949}   & \textbf{7949}   & 1.3         & 8.7       & 0.0    & 0.2    \\
 20   & 25  & B4  & 3   & 20  & 7376  & \textbf{7689}   & \textbf{7689}   & 0.0         & 4.1       & 0.0    & 0.0    \\
 21   & 25  & B4  & 3   & 50  & 7370  & \textbf{7615}   & \textbf{7615}   & 0.6         & 3.2       & 0.0    & 0.1    \\
 22   & 25  & B4  & 5   & 10  & 9678  & 11361           & \textbf{11125}  & 0.8         & 13.0      & -2.1   & -1.9   \\
 23   & 25  & B4  & 5   & 20  & 9139  & 11130           & \textbf{10779}  & 0.6         & 15.2      & -3.3   & -3.0   \\
 24   & 25  & B4  & 5   & 50  & 8999  & 11003           & \textbf{10608}  & 7.8         & 15.2      & -3.7   & -3.2   \\
\hline

 25   & 50  & B1  & 3   & 10  & 10359 & 25237           & \textbf{24117}  & 6.1         & 57.0      & -4.6   & -4.1   \\
 26   & 50  & B1  & 3   & 20  & 9406  & 24455           & \textbf{22783}  & 6.3         & 58.7      & -7.3   & -6.8   \\
 27   & 50  & B1  & 3   & 50  & 9950  & 23951           & \textbf{22627}  & 16.2        & 56.0      & -5.9   & -4.7   \\
 28   & 50  & B1  & 5   & 10  & 14607 & 32491           & \textbf{31351}  & 14.0        & 53.4      & -3.6   & -0.7   \\
 29   & 50  & B1  & 5   & 20  & 14661 & 32915           & \textbf{31908}  & 6.7         & 54.1      & -3.2   & -2.3   \\
 30   & 50  & B1  & 5   & 50  & 14791 & 32203           & \textbf{31127}  & 47.6        & 52.5      & -3.5   & -2.0   \\
 31   & 50  & B2  & 3   & 10  & 6488  & 20966           & \textbf{20244}  & 22.2        & 68.0      & -3.6   & -2.9   \\
 32   & 50  & B2  & 3   & 20  & 5821  & 18854           & \textbf{18449}  & 3.7         & 68.4      & -2.2   & -1.6   \\
 33   & 50  & B2  & 3   & 50  & 6423  & 18876           & \textbf{18449}  & 3.0         & 65.2      & -2.3   & -1.8   \\
 34   & 50  & B2  & 5   & 10  & 10192 & 25377           & \textbf{25067}  & 12.3        & 59.3      & -1.2   & -0.4   \\
 35   & 50  & B2  & 5   & 20  & 9250  & 25977           & \textbf{24777}  & 19.2        & 62.7      & -4.8   & -3.5   \\
 36   & 50  & B2  & 5   & 50  & 8858  & 24771           & \textbf{24117}  & 25.0        & 63.3      & -2.7   & -1.7   \\
 37   & 50  & B3  & 3   & 10  & 15323 & 31508           & \textbf{30858}  & 35.2        & 50.3      & -2.1   & -0.8   \\
 38   & 50  & B3  & 3   & 20  & 15323 & 31505           & \textbf{30932}  & 7.8         & 50.5      & -1.9   & -0.5   \\
 39   & 50  & B3  & 3   & 50  & 14984 & 31299           & \textbf{30672}  & 25.4        & 51.1      & -2.0   & -0.7   \\
 40   & 50  & B3  & 5   & 10  & 25085 & 45507           & \textbf{44289}  & 35.8        & 43.4      & -2.8   & -1.3   \\
 41   & 50  & B3  & 5   & 20  & 24434 & 45670           & \textbf{44806}  & 38.9        & 45.5      & -1.9   & -0.4   \\
 42   & 50  & B3  & 5   & 50  & 22912 & 43860           & \textbf{42848}  & 44.6        & 46.5      & -2.4   & 0.6    \\
 43   & 50  & B4  & 3   & 10  & 10814 & 26210           & \textbf{25393}  & 7.1         & 57.4      & -3.2   & -2.3   \\
 44   & 50  & B4  & 3   & 20  & 10919 & 25813           & \textbf{24574}  & 53.5        & 55.6      & -5.0   & -4.2   \\
 45   & 50  & B4  & 3   & 50  & 11238 & 25647           & \textbf{24460}  & 36.9        & 54.1      & -4.9   & -4.1   \\
 46   & 50  & B4  & 5   & 10  & 15175 & 32768           & \textbf{31372}  & 40.1        & 51.6      & -4.4   & -2.9   \\
 47   & 50  & B4  & 5   & 20  & 14973 & 32833           & \textbf{31658}  & 14.2        & 52.7      & -3.7   & -2.2   \\
 48   & 50  & B4  & 5   & 50  & 14630 & 32316           & \textbf{31366}  & 47.3        & 53.4      & -3.0   & -1.5   \\
\hline

 49   & 100 & B1  & 3   & 10  & 17217 & 100173          & \textbf{93843}  & 148.5       & 81.7      & -6.7   & -5.9   \\
 50   & 100 & B1  & 3   & 20  & 17120 & 97933           & \textbf{90707}  & 124.6       & 81.1      & -8.0   & -6.9   \\
 51   & 100 & B1  & 3   & 50  & 17111 & 97922           & \textbf{90642}  & 117.3       & 81.1      & -8.0   & -7.3   \\
 52   & 100 & B1  & 5   & 10  & 27533 & 116348          & \textbf{110271} & 135.5       & 75.0      & -5.5   & -4.0   \\
 53   & 100 & B1  & 5   & 20  & 27237 & 116494          & \textbf{109971} & 77.4        & 75.2      & -5.9   & -4.6   \\
 54   & 100 & B1  & 5   & 50  & 27189 & 115508          & \textbf{108961} & 137.4       & 75.0      & -6.0   & -5.3   \\
 55   & 100 & B2  & 3   & 10  & 10218 & 92689           & \textbf{85734}  & 120.9       & 88.1      & -8.1   & -7.4   \\
 56   & 100 & B2  & 3   & 20  & 9571  & 88579           & \textbf{81464}  & 115.6       & 88.3      & -8.7   & -7.7   \\
 57   & 100 & B2  & 3   & 50  & 9296  & 87901           & \textbf{81492}  & 159.7       & 88.6      & -7.9   & -7.0   \\
 58   & 100 & B2  & 5   & 10  & 16074 & 102688          & \textbf{95578}  & 77.2        & 83.2      & -7.4   & -6.3   \\
 59   & 100 & B2  & 5   & 20  & 15631 & 101898          & \textbf{96129}  & 91.8        & 83.7      & -6.0   & -5.0   \\
 60   & 100 & B2  & 5   & 50  & 15104 & 99941           & \textbf{93800}  & 76.6        & 83.9      & -6.5   & -5.5   \\
 61   & 100 & B3  & 3   & 10  & 26205 & 111116          & \textbf{108917} & 68.4        & 75.9      & -2.0   & -1.1   \\
 62   & 100 & B3  & 3   & 20  & 26135 & 111494          & \textbf{109436} & 131.7       & 76.1      & -1.9   & -0.9   \\
 63   & 100 & B3  & 3   & 50  & 25066 & 110730          & \textbf{108655} & 65.5        & 76.9      & -1.9   & -0.7   \\
 64   & 100 & B3  & 5   & 10  & 46806 & 141233          & \textbf{136503} & 82.6        & 65.7      & -3.5   & -2.5   \\
 65   & 100 & B3  & 5   & 20  & 45595 & 141348          & \textbf{137609} & 144.0       & 66.9      & -2.7   & -1.7   \\
 66   & 100 & B3  & 5   & 50  & 43378 & 140308          & \textbf{135262} & 151.1       & 67.9      & -3.7   & -2.6   \\
 67   & 100 & B4  & 3   & 10  & 18905 & 100798          & \textbf{96329}  & 81.8        & 80.4      & -4.6   & -3.6   \\
 68   & 100 & B4  & 3   & 20  & 18121 & 100817          & \textbf{94064}  & 66.6        & 80.7      & -7.2   & -6.1   \\
 69   & 100 & B4  & 3   & 50  & 17810 & 101713          & \textbf{93914}  & 155.7       & 81.0      & -8.3   & -7.5   \\
 70   & 100 & B4  & 5   & 10  & 26710 & 115442          & \textbf{108894} & 48.0        & 75.5      & -6.0   & -4.3   \\
 71   & 100 & B4  & 5   & 20  & 26570 & 115037          & \textbf{109064} & 174.9       & 75.6      & -5.5   & -4.3   \\
 72   & 100 & B4  & 5   & 50  & 26330 & 114366          & \textbf{108810} & 101.6       & 75.8      & -5.1   & -4.1   \\

\hline
 73   & 200 & B1  & 3   & 10  & 30470 & 375104          & \textbf{363104} & 218.5       & 91.6      & -3.3   & -3.0   \\
 74   & 200 & B1  & 3   & 20  & 30470 & 374153          & \textbf{356945} & 237.8       & 91.5      & -4.8   & -4.4   \\
 75   & 200 & B1  & 3   & 50  & 30470 & 375918          & \textbf{355549} & 261.0       & 91.4      & -5.7   & -5.1   \\
 76   & 200 & B1  & 5   & 10  & 50565 & 409442          & \textbf{397386} & 239.3       & 87.3      & -3.0   & -2.5   \\
 77   & 200 & B1  & 5   & 20  & 50565 & 415919          & \textbf{394957} & 113.4       & 87.2      & -5.3   & -4.3   \\
 78   & 200 & B1  & 5   & 50  & 50565 & 421503          & \textbf{392877} & 179.9       & 87.1      & -7.3   & -6.8   \\
 79   & 200 & B2  & 3   & 10  & 16107 & 360694          & \textbf{346867} & 325.6       & 95.4      & -4.0   & -3.7   \\
 80   & 200 & B2  & 3   & 20  & 14460 & 350099          & \textbf{338312} & 250.0       & 95.7      & -3.5   & -3.1   \\
 81   & 200 & B2  & 3   & 50  & 14442 & 344132          & \textbf{338478} & 229.3       & 95.7      & -1.7   & -1.3   \\
 82   & 200 & B2  & 5   & 10  & 27553 & 382758          & \textbf{368084} & 163.5       & 92.5      & -4.0   & -3.5   \\
 83   & 200 & B2  & 5   & 20  & 26310 & 387238          & \textbf{365656} & 257.3       & 92.8      & -5.9   & -5.4   \\
 84   & 200 & B2  & 5   & 50  & 25987 & 384555          & \textbf{363227} & 220.2       & 92.8      & -5.9   & -5.5   \\
 85   & 200 & B3  & 3   & 10  & 45272 & \textbf{393751} & 393825          & 249.9       & 88.5      & 0.0    & 0.5    \\
 86   & 200 & B3  & 3   & 20  & 45229 & 394673          & \textbf{394159} & 196.2       & 88.5      & -0.1   & 0.6    \\
 87   & 200 & B3  & 3   & 50  & 44800 & 394898          & \textbf{393288} & 218.3       & 88.6      & -0.4   & 0.4    \\
 88   & 200 & B3  & 5   & 10  & 85028 & 457451          & \textbf{449657} & 283.8       & 81.1      & -1.7   & -1.3   \\
 89   & 200 & B3  & 5   & 20  & 85028 & 459525          & \textbf{451109} & 231.0       & 81.2      & -1.9   & -1.0   \\
 90   & 200 & B3  & 5   & 50  & 79786 & 464271          & \textbf{446762} & 167.3       & 82.1      & -3.9   & -3.0   \\
 91   & 200 & B4  & 3   & 10  & 31557 & 373626          & \textbf{368836} & 287.4       & 91.4      & -1.3   & -1.1   \\
 92   & 200 & B4  & 3   & 20  & 31557 & 374781          & \textbf{364081} & 249.2       & 91.3      & -2.9   & -2.2   \\
 93   & 200 & B4  & 3   & 50  & 31491 & 371865          & \textbf{364175} & 345.9       & 91.4      & -2.1   & -1.8   \\
 94   & 200 & B4  & 5   & 10  & 49358 & 408022          & \textbf{394708} & 120.9       & 87.5      & -3.4   & -2.7   \\
 95   & 200 & B4  & 5   & 20  & 49196 & 406270          & \textbf{394761} & 244.5       & 87.5      & -2.9   & -2.4   \\
 96   & 200 & B4  & 5   & 50  & 48658 & 412571          & \textbf{394668} & 244.4       & 87.7      & -4.5   & -4.3   \\
\hline

\end{longtable}}}


\bibliographystyle{elsarticle-harv} 
\bibliography{references_v3}

\end{document}